\documentclass[pdflatex,sn-mathphys-num]{sn-jnl}

\usepackage{graphicx}%
\usepackage{multirow}%
\usepackage{amsmath,amssymb,amsfonts}%
\usepackage{amsthm}%
\usepackage{mathrsfs}%
\usepackage[title]{appendix}%
\usepackage{xcolor}%
\usepackage{textcomp}%
\usepackage{manyfoot}%
\usepackage{booktabs}%
\usepackage{algorithm}%
\usepackage{algorithmicx}%
\usepackage{algpseudocode}%
\usepackage{listings}%
\usepackage{subcaption}
\usepackage{tabularx}

\theoremstyle{thmstyleone}%
\theoremstyle{thmstyletwo}%

\theoremstyle{thmstylethree}%

\begin{document}

\title[Article Title]{Advancing Symbolic Integration in Large Language Models: Beyond Conventional Neurosymbolic AI}

\author*[1]{\fnm{Maneeha} \sur{Rani}}\email{m.rani3-2022@hull.ac.uk}
\author[1]{\fnm{Bhupesh Kumar} \sur{Mishra}}\email{bhupesh.mishra@hull.ac.uk}
\author[2]{\fnm{Dhavalkumar} \sur{Thakker}}\email{D.Thakker@hull.ac.uk}

\affil*[1]{\orgdiv{Data Science, AI \& Modelling Centre (DAIM)}, \orgdiv{School of Digital and Physical Sciences},
  \orgname{University of Hull},
  \orgaddress{\street{Cottingham Road}, \city{Hull}, \country{UK}}}

\affil[2]{\orgdiv{School of Digital and Physical Sciences},
  \orgname{University of Hull},
  \orgaddress{\street{Cottingham Road}, \city{Hull}, \country{UK}}}


\abstract{LLMs have demonstrated highly effective learning, human-like response generation, and decision-making capabilities in high-risk sectors. However, these models remain black boxes because they struggle to ensure transparency in responses. The literature has explored numerous approaches to address transparency challenges in LLMs, including Neurosymbolic AI (NeSy AI). NeSy AI approaches were primarily developed for conventional neural networks and are not well-suited to the unique features of LLMs. Consequently, there is a limited systematic understanding of how symbolic AI can be effectively integrated into LLMs. This paper aims to address this gap by first reviewing established NeSy AI methods and then proposing a novel taxonomy of symbolic integration in LLMs, along with a roadmap to merge symbolic techniques with LLMs. The roadmap introduces a new categorisation framework across four dimensions by organising existing literature within these categories. These include symbolic integration across various stages of LLM, coupling mechanisms, architectural paradigms, as well as algorithmic and application-level perspectives. The paper thoroughly identifies current benchmarks, cutting-edge advancements, and critical gaps within the field to propose a roadmap for future research. By highlighting the latest developments and notable gaps in the literature, it offers practical insights for implementing frameworks for symbolic integration into LLMs to enhance transparency.
}



\maketitle
 
\section{Introduction}
\label{sec1}
Recently, Language Models (LMs) have gained significant attention due to their remarkable performance and potential applications across numerous fields [1]. A few popular LMs include Llama3 [2], GPT, Falcon [3], MedPalm2 [4], BioBERT [5], ClinicalBERT [6], BlueBERT [7], MedNLI [8], and Gemini [9]. Despite LLMs' advanced capabilities, it is evident that further improvements are necessary to address the specific requirements of various industries. LLMs are quite effective at language generation; however, ensuring transparency in responses remains a notable limitation. Several critical factors undermine transparency in LLMs, such as hallucination [10], non-robustness [11], trustworthiness [12], fairness [13], biases [14], security and privacy concerns [15], lack of interpretability [16] and explainability [17]. Considering these factors, expecting complete transparency in response generation, solely from a transformer-based architecture seems unrealistic. [18]. This underscores the urgency and importance of our research in this field.

Traditional neural networks faced challenges similar to those encountered by LLMs today [18]. In literature, NeSy approaches have demonstrated potential in overcoming some of these challenges by combining neural learning with symbolic AI [19]. NeSy AI offers several capabilities, including learning from relatively fewer data [20] compared to substantial training data to train the transformer architecture [21], offering reduced computational complexity [19]. Further, NeSy AI can handle data not seen during training, improving a model's ability to generalise beyond familiar cases. This out-of-distribution data handling can enhance decision-making by precisely interpreting unexpected results, particularly in critical areas like healthcare. These capabilities are primarily ensured by the symbolic component, which provides the logical foundation necessary for these functions [20]. The symbolic component of NeSy AI architectures can be implemented through various paradigms, including Knowledge Graphs (KGs), rule-based engines or logic-based systems. Figure 1 illustrates these paradigms, where each implementation offers distinct advantages for structured reasoning and knowledge representation to leverage both learned patterns and explicit symbolic knowledge. 
\begin{figure}[t]
\centering
\includegraphics[width=1\textwidth]{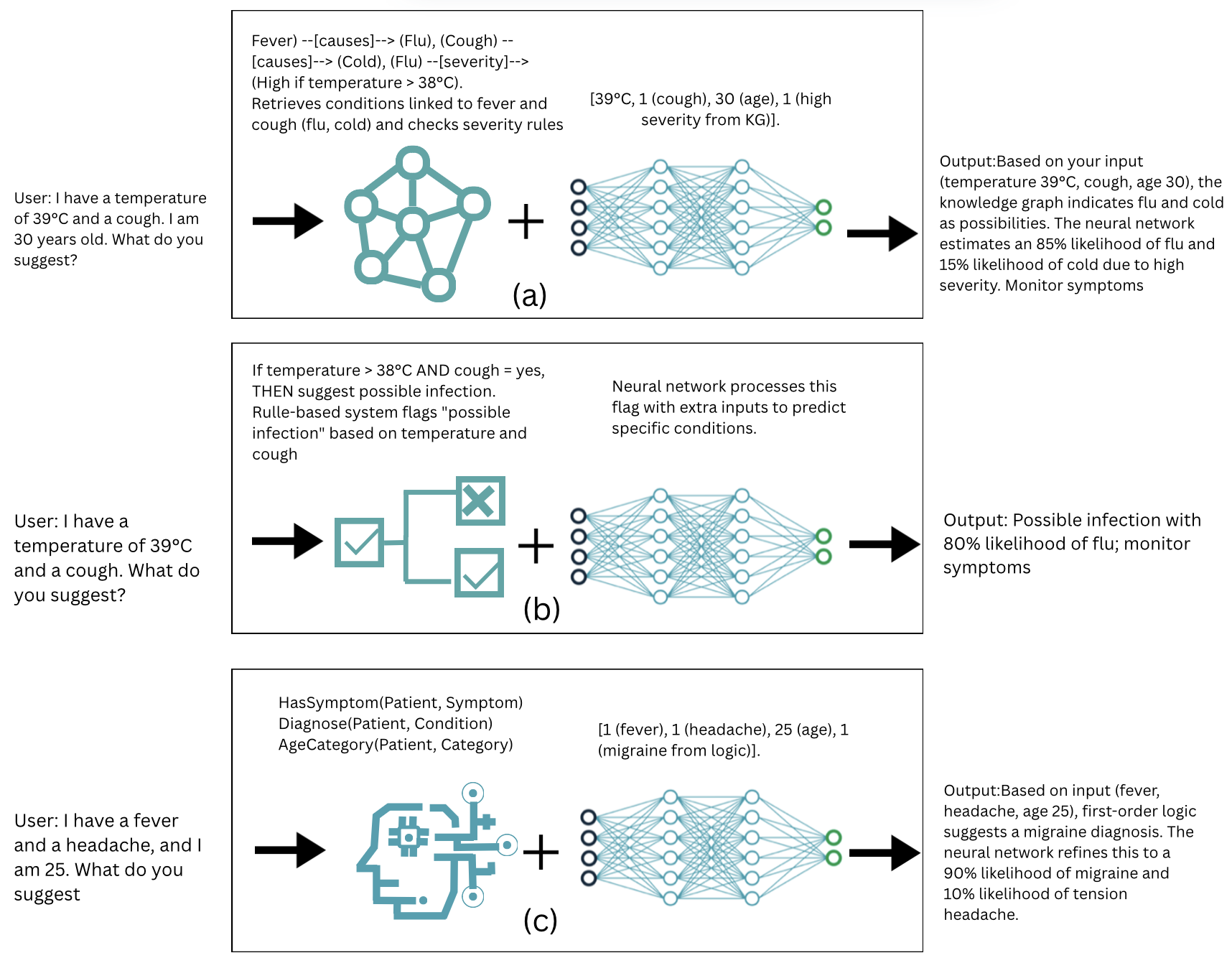} 
\caption{Neurosymbolic AI architectures with different symbolic reasoning paradigms: (a) Knowledge Graph-Neural Network Integration; (b) Rule-Neural Network Integration and (c) Logic-Neural Network Integration.  Each subfigure shows the information flow between symbolic and neural components with inputs and outputs at each processing stage.}\label{fig1}
\end{figure}

By integrating a symbolic component with LLMs, LLMs can acquire structured knowledge, which in turn allows them to perform logical reasoning, explainability, and interpretability. However, LLMs differ fundamentally from traditional neural networks because of their large parameter scale, language modelling abilities, autoregressive token-based generation and capacity to produce context-dependent outputs. Therefore, using existing NeSy AI frameworks on LLMs may not fully capture the unique properties and operational requirements of LLMs. These frameworks may require adaptation to accommodate the distinctive characteristics of LLMs. Instead, Symbolic integration with LLMs can be seen as a specialised type of NeSy AI that requires customised integration strategies, given the distinct features of LLMs. Furthermore, integrating symbolic AI poses its own challenges, including effective knowledge representation and reasoning over large-scale language contexts. This paper addresses this gap by first reviewing established NeSy AI approaches and frameworks and then proposing a roadmap comprised of a novel taxonomy for symbolic integration in LLMs. The main contributions of this review work include:

\begin{enumerate}[1.]

\item Categorisation: This article introduces a novel categorisation for combining symbolic AI with LLMs, organising approaches into categories based on integration stages in LLM development. It examines coupling levels, architectural paradigms, and frameworks at the application and algorithm levels.
\item Benchmarks: This paper critically examines the benchmarks used in the literature to evaluate methodologies that integrate symbolic AI with LLMs to improve reasoning and interpretability in LLM-generated responses. 
\item Comparative analysis to address explainability: This paper examines current applications of symbolic integrated LLMs across different fields, such as reasoning, interpretability, planning, and explainability. It investigates how combining symbolic AI with LLMs can enhance explainability on numerous levels and provides a comparative analysis of existing methods, highlighting their limitations and potential future developments. 
\item State-of-the-art achievements and Challenges: This article highlights state-of-the-art accomplishments and challenges in existing symbolic integrated LLMs research, with promising future research directions. 
\end{enumerate}

The rest of the paper is organised as follows: Section 2 explores structured review methodology. Section 3 explains NeSy AI, including various approaches and frameworks. Section 4 introduces the integration of LLMs with symbolic AI across different phases, including the coupling, application, and algorithm-level perspectives. Section 5 presents architectural paradigms. Section 6 discusses benchmarks for KG integrated LLMs and logic integrated LLMs alongwith the challenges. Section 7 integration role and explains explainability at multiple levels of LLMs and provides a comparative analysis of approaches targeting explainability in LLMs using symbolic AI. Section 8 discusses various state-of-the-art achievements and challenges.

\section{Structured Literature Review Methodology}

\label{sec2}

This study employs a systematic literature review (SLR) methodology to ensure comprehensive coverage of existing literature while maintaining methodological rigour. The review focused on addressing the following research questions and keywords:
\begin{itemize}
    \item \textbf{RQ1:} What are the state-of-the-art methods for integrating symbolic AI with LLMs?
    \item \textbf{RQ2:} In what ways do these integrations enhance explainability, reasoning, and trustworthiness of LLMs?
    \item \textbf{RQ3:} What integration stages, coupling strategies, and architectural paradigms have been proposed?
    \item \textbf{RQ4:} What evaluation practices, benchmarks, and datasets are used to assess symbolic integration with LLMs?
    \item \textbf{RQ5:} What are the key achievements, open challenges, and future research directions in this domain?
\end{itemize}

The search strategy was developed by identifying a set of precise keywords that reflect the core dimensions of this review. These included: 'Symbolic AI', 'Large Language Models (LLM),' 'Symbolic Integration', 'Knowledge Graph (KG) Integration', 'Logic Integration', 'Explainability', 'Reasoning', 'Benchmarks', 'Challenges' and 'Achievements'.
From these keywords, search components were derived to capture specific aspects of LLM–symbolic integration. The search components included 'Symbolic integration with LLM', 'KG OR logic integration with LLM', 'Symbolic integration with LLM for explainability', 'KG OR logic integration AND explainability', 'Symbolic reasoning AND LLM', 'KG integration in the pre-training stage', 'KG integration in the fine-tuning stage', and 'KG integration in the post-training stage'. These combinations allowed the search to capture integration stages, coupling information, algorithm-level, application-level and architecture-level integration details, benchmarks, integration role  as well as state-of-the-art achievements and challenges. To ensure precision and reduce irrelevant retrievals, the search queries were adapted to the indexing requirements of each database. Eligibility of studies was determined by applying inclusion and exclusion criteria tailored to the role of symbolic integration in LLMs.

\begin{table}[h]
\scriptsize
\setlength{\tabcolsep}{4pt} 
\renewcommand{\arraystretch}{1.2} 

\caption{Search keywords along with Inclusion and Exclusion criteria}
\label{tab:criteria}

\begin{tabular}{@{}p{2.5cm}p{5.5cm}p{6cm}@{}}
\toprule
\textbf{Conceptual Category} & \textbf{Inclusion Criteria} & \textbf{Exclusion Criteria} \\
\midrule
Symbolic AI and LLMs & Integration of Symbolic AI with LLMs to address challenges of LLM & Focusing solely on traditional neural networks without LLM components \\

NeSy approaches & NeSy AI frameworks and approaches addressing neural challenges & NeSy AI frameworks and approaches focusing on symbolic AI \\

Symbolic AI integration at several stages of LLM & Symbolic AI integration at various stages of LLM including pre-training, inference, post-training, fine-tuning & Studies not explicitly specifying the integration stage of LLM or lacking methodologies for such integration \\

NeSy coupling & Symbolic integration interaction level/ tight /loose coupling & Integration with LLMs at several coupling levels \\

Algorithm level symbolic integration with LLMs & Symbolic logic and KG-based integration to enhance LLM responses at algorithm level & Other symbolic approaches integration at algorithm level without based integration to enhance LLM responses \\

Application level & Symbolic logic and KG-based integration to enhance LLM responses at application level & Other symbolic approaches integration at algorithm level without based integration to enhance LLM responses \\

LLM to symbolic flow & Focused only on information flow from logic or formal methods conversion, action schema generation to enhance LLMs & Studies that do not involve LLM-generated data feeding into symbolic systems or lack methodological details on this integration flow \\

Symbolic to LLM & Papers demonstrating enhancements in explainability, reasoning, or interpretability through symbolic AI & Research lacking information flow direction or specific integration methodologies \\

Symbolic integration with LLMs Hybrid Models & Studies focusing on hybrid models combining symbolic AI and LLMs to enhance reasoning, interpretability, or task performance & Studies that do not explicitly address hybrid architectures or lack detailed integration frameworks of symbolic AI with LLMs \\

Symbolic Integration Benchmarks & Knowledge Graphs (KGs) and Symbolic Logic Benchmarks to address interpretability and reasoning & Studies that do not provide or utilize specific benchmarks, datasets, or evaluation metrics for symbolic integration with LLMs are excluded \\

The role of symbolic AI integration & Studies examining the role of symbolic AI integration in enhancing LLM capabilities such as reasoning, explainability & Studies that do not address the impact or role of symbolic AI in LLM functionality or lack practical insights into its applications \\

State-of-the-art challenges and achievements & Studies highlighting state-of-the-art challenges and achievements in symbolic AI integration with LLMs & Studies that do not address current challenges or notable advancements in symbolic AI and LLM integration \\

\botrule
\end{tabular}
\end{table}
\normalsize
Table ~\ref{tab:criteria}  summarises these criteria, categorised according to conceptual components of integration.This review considered studies published between 2018 and February 2025. The relatively low number of studies from early 2025 is attributed to the limited availability of articles at the time of data collection. This study employs the PRISMA 2020 guidelines to conduct a systematic literature review (SLR). Extensive searches were conducted across Scopus, IEEE Xplore, ACM Digital Library, SpringerLink, Elsevier (ScienceDirect), PubMed, MDPI, ACL Anthology, AAAI, IJCAI, OpenReview, and ArXiv. A few high-impact works are available in arXiv, therefore the inclusion of arXiv papers was deemed necessary to ensure comprehensive coverage of recent and influential contributions. For this review, a three-rule selection strategy was applied to arXiv papers. Publications were included if they satisfied at least two of the following criteria: (i) published in 2023 or 2024, (ii) a citation count greater than 30, and (iii) demonstrated high topical relevance.  In addition, Google Scholar was used as a supplementary source to capture grey literature and cross-check coverage. Records from Google Scholar were subsequently mapped to the corresponding databases to avoid duplication.

\begin{figure}[H]
\centering
\includegraphics[width=0.60\textwidth]{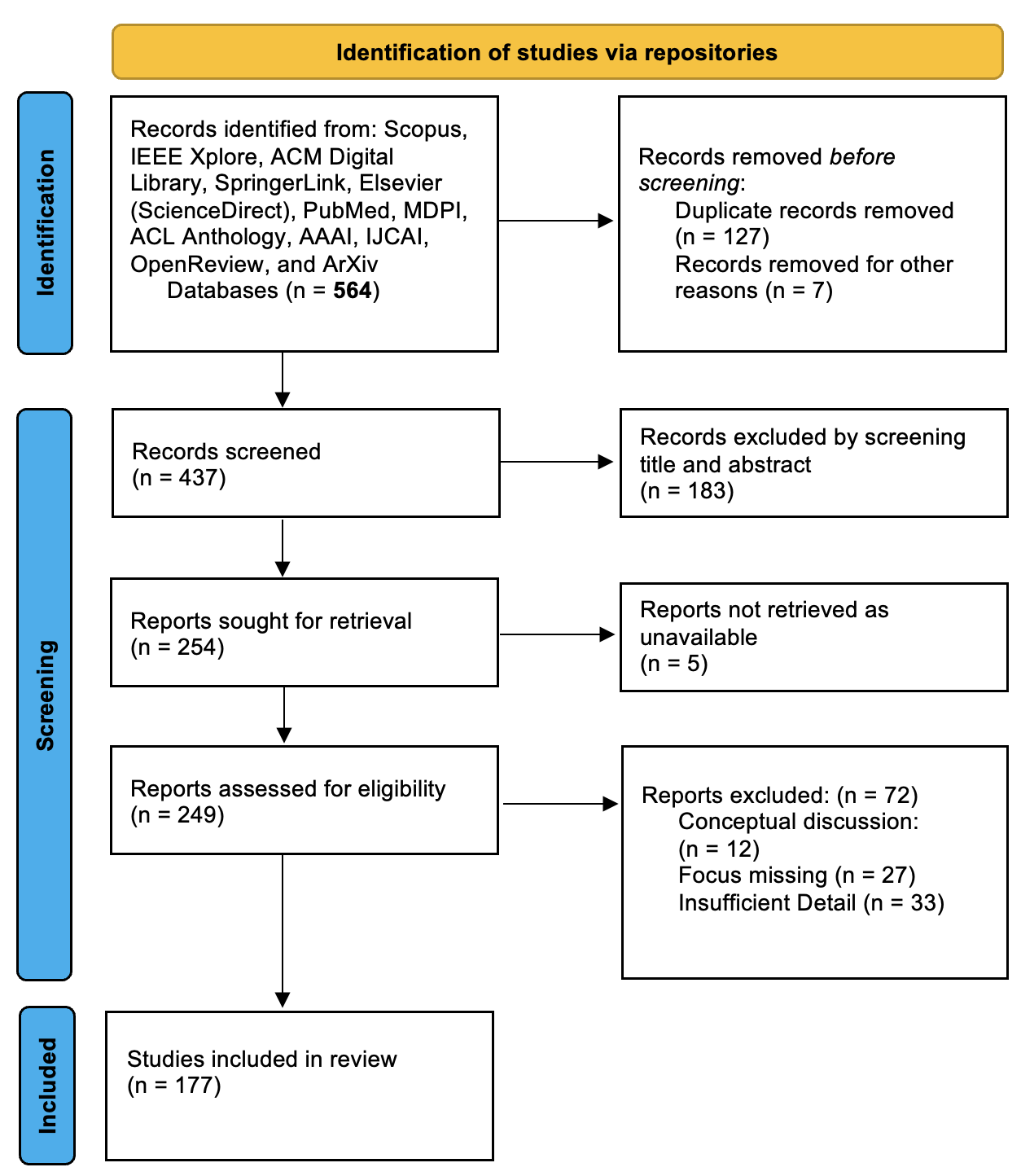}
\caption{PRISMA flow diagram showing the systematic literature review process}
\label{fig:prisma}  
\end{figure}
\begin{figure}[htbp]
    \centering
    \begin{subfigure}{0.45\textwidth}
        \centering
        \includegraphics[width=\textwidth]{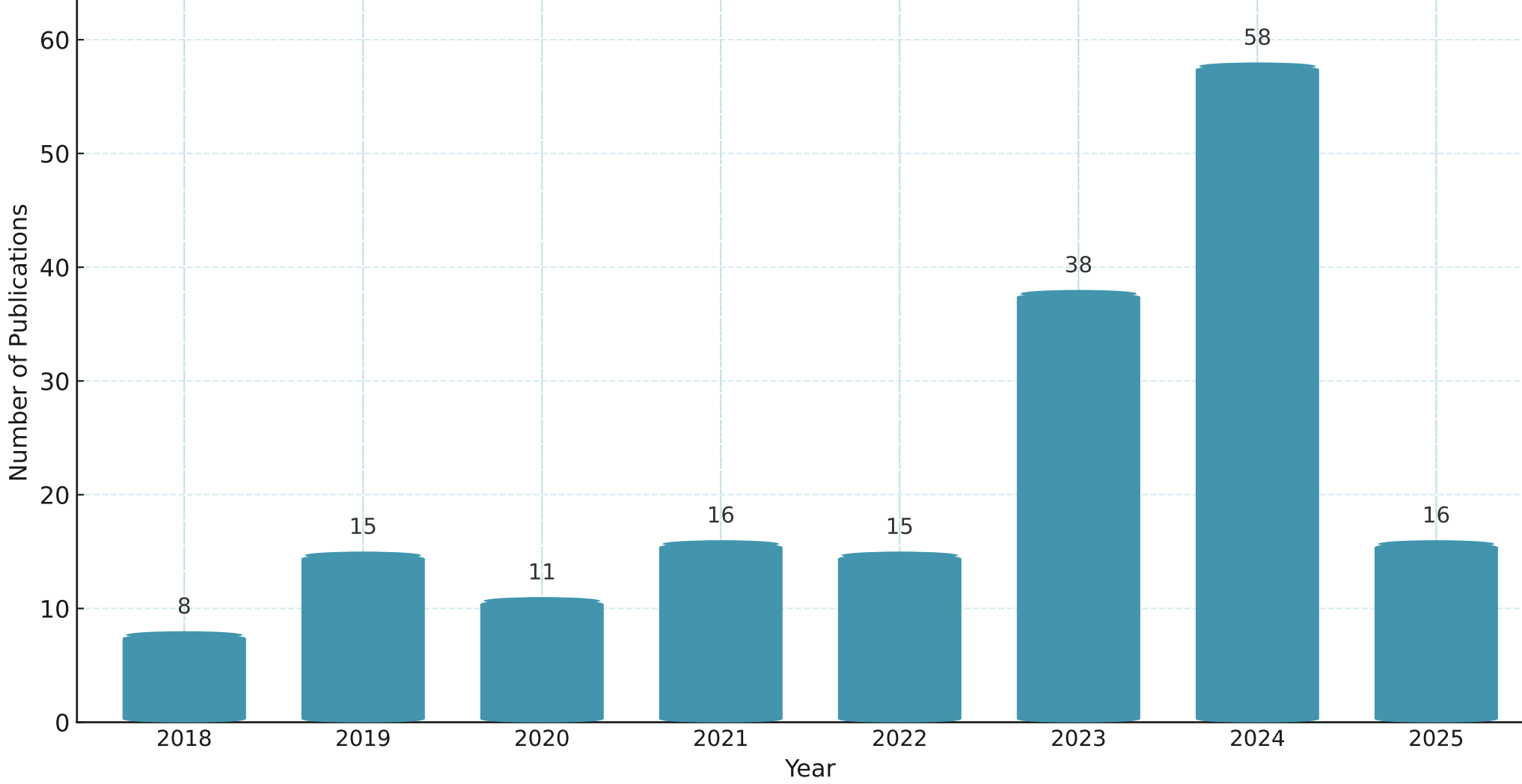}
        \caption{Database distribution of selected studies}%
        \label{fig:databases}
    \end{subfigure}
    \hfill
    \begin{subfigure}{0.45\textwidth}
        \centering
        \includegraphics[width=\textwidth]{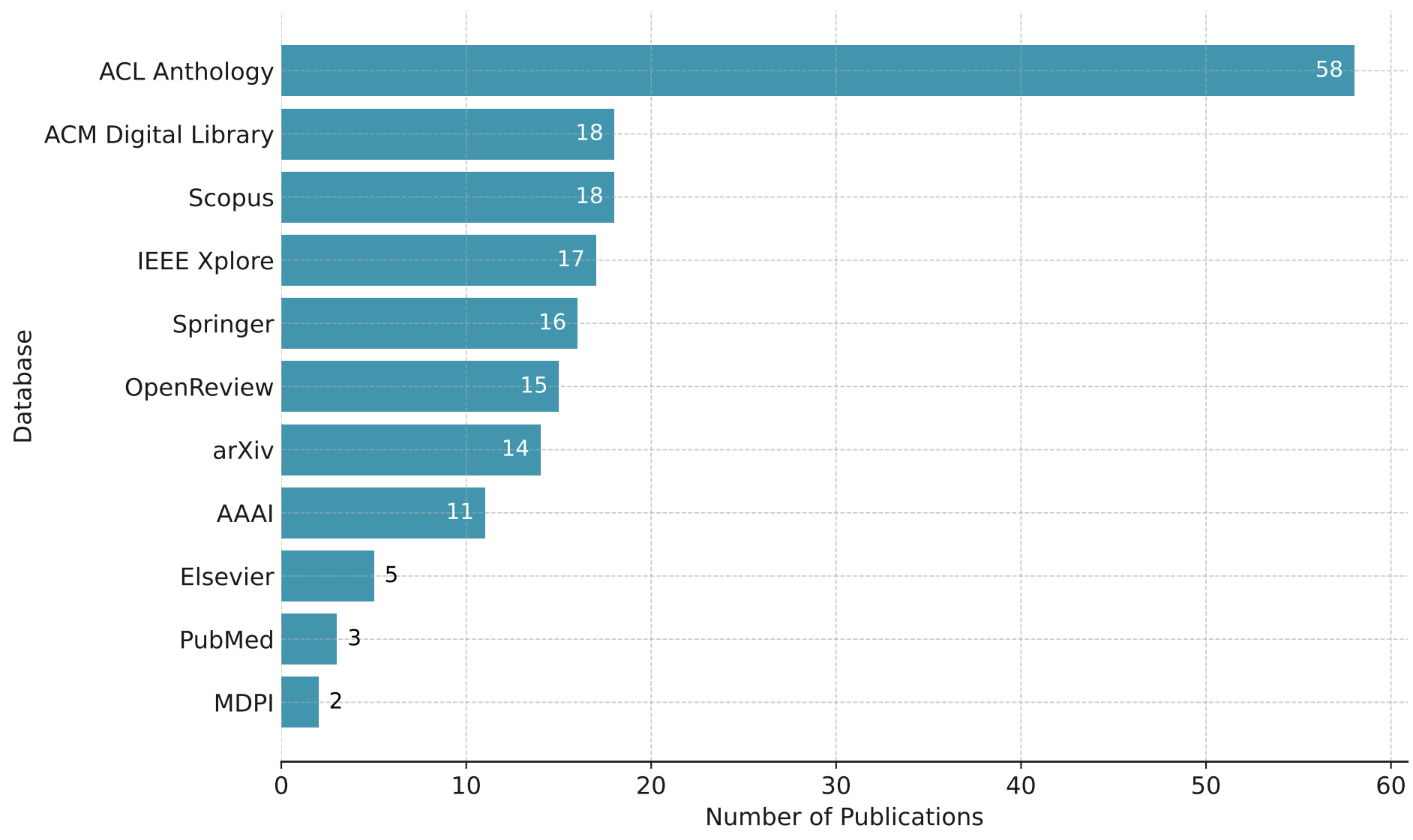}
        \caption{Temporal distribution of studies by year}%
        \label{fig:years}
    \end{subfigure}

    \vspace{0.75em}

    \begin{subfigure}{0.45\textwidth}
        \centering
        \includegraphics[height=5cm]{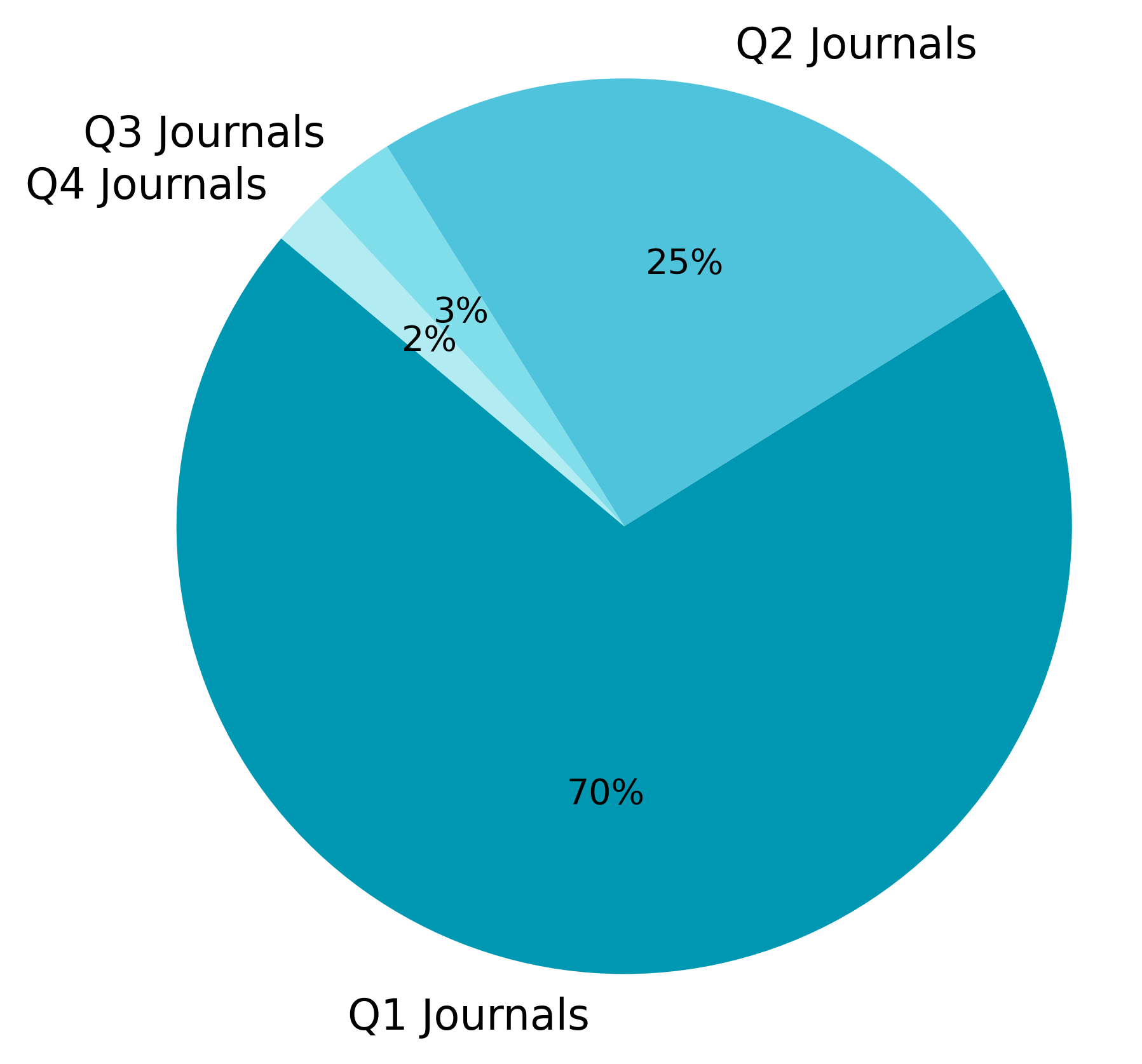}
        \caption{Journal quartiles }%
        \label{fig:journals_quartiles}
    \end{subfigure}
    \hfill
    \begin{subfigure}{0.45\textwidth}
        \centering
        \includegraphics[height=5cm]{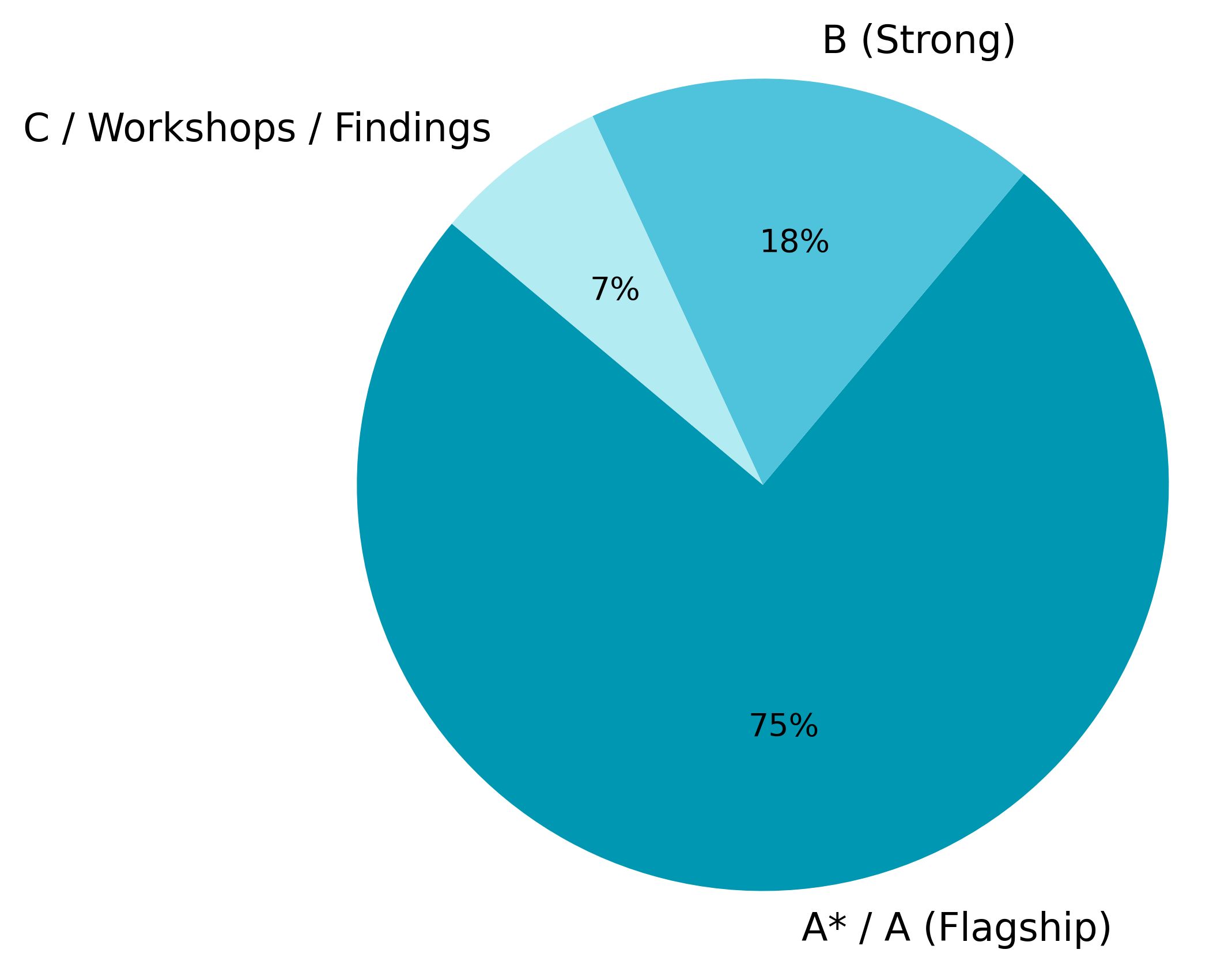}
        \caption{Conference tiers }%
        \label{fig:conferences_tiers}
    \end{subfigure}

    \caption{Distribution analysis of included studies in the systematic literature review: 
    (a) database coverage, (b) temporal spread by year, (c) journal quality distribution across quartiles, and (d) conference quality distribution across tier groups.}
    \label{fig:distribution}
\end{figure}
The initial database search returned a total of 564 records. After removing 127 duplicates, 437 records remained for screening. Titles and abstracts were screened to exclude 183 clearly irrelevant papers. Full text is extracted, which resulted in 5 reports being unavailable; therefore, a full-text eligibility assessment was conducted on these 249 papers. During this stage, 72 papers were excluded because of conceptual-level discussion or opinion, a focus outside symbolic integration with LLMs and insufficient methodological detail. Finally, 177 studies were included in the review for qualitative synthesis. These studies form the evidence for the analysis presented in this paper. The venue analysis indicates that ACL Anthology contributed the most significant number of publications. Year-wise trends highlight a sharp rise in publications from 2019 onwards, peaking in 2023–2024. The detailed methodology for study identification, screening, and inclusion is presented in PRISMA guidelines (Figure~\ref{fig:prisma}).The distribution of the included studies across publication venues and publication years is shown in Figures~\ref{fig:databases} and~\ref{fig:years}. In addition to the temporal and database coverage (Figs. \ref{fig:databases}–\ref{fig:years}), the venue quality overview (Figs. \ref{fig:journals_quartiles}–\ref{fig:conferences_tiers}) indicates that journal publications predominantly appear in Q1 outlets, with a smaller share in Q2 and negligible Q3/Q4 presence. The conference set is similarly concentrated in flagship A/A venues*, with modest representation in B-tier and limited contributions from workshops/other tracks. This distribution reflects an emphasis on high-quality venues across both journals and conferences.

For each study, we extracted details on the integration method, symbolic component, coupling degree, application vs algorithmic analysis, architectural paradigms, integration role, application domain, transparency aspects, and identified challenges. A narrative synthesis was conducted to systematically group and classify the included literature, enabling the identification of thematic patterns, methodological approaches, and conceptual frameworks across studies. A novel taxonomy spanning four dimensions is designed, including integration stages, coupling strategies, algorithmic versus application-level approaches and architectural paradigms. This structured methodology ensures comprehensiveness while directly addressing the research gap on symbolic integration in LLMs. It also provides a rigorous foundation for the proposed categorisation and subsequent analysis of challenges and future research directions.

\begin{figure}[t]
\centering
\includegraphics[width=0.85\textwidth]{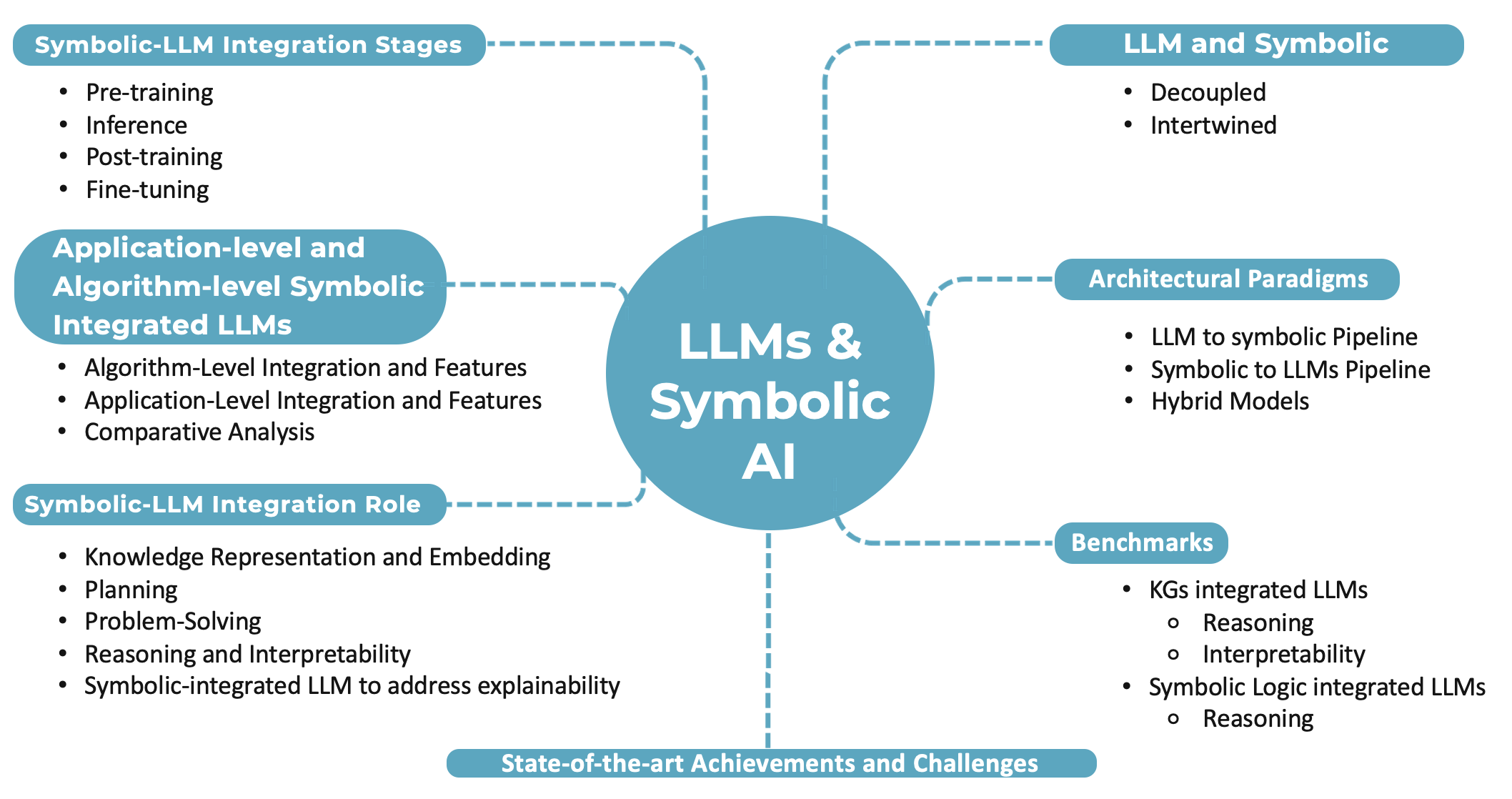}
\caption{Roadmap for Symbolic Integration with Large Language Models (LLMs)}\label{fig:roadmap}
\end{figure}

Figure~\ref{fig:roadmap} offers a detailed overview of the main sections of this review paper. It demonstrates how symbolic integration with LLMs can be achieved by exploring different integration perspectives. It acts as a roadmap for readers, guiding them through the sequence of topics in the upcoming sections.

\begin{table}[htbp]
\scriptsize 
\renewcommand{\arraystretch}{1.3} 
\setlength{\tabcolsep}{6pt} 
\caption{Neurosymbolic Categories and Approaches}\label{tab:neurocat}
\begin{tabular}{p{2cm}p{2cm}p{4cm}p{4cm}}
\toprule
\textbf{Category} & \textbf{Approaches} & \textbf{Description} & \textbf{Example Tasks} \\
\midrule
High reasoning and low learning systems &
Probabilistic Logic Programming (ProbLog) &
A probabilistic programming language to define probability distribution and truth values generation by reasoning mechanism [24] [25]. &
More focused on machine learning; can perform marginal inference; capable of modelling and reasoning about uncertain domains [23]. \\
& Markov Logic Network (MLN) &
Applies the idea of Markov networks to logical problems and can generate inferences by learning logical rules. & -- \\
& Inductive Logic Programming (ILP) &
A classical rule-based approach using logic programming to represent knowledge. &
Capable of providing self-explanatory symbolic, goal-driven, and causal interpretation of data [26]\\
\midrule
Low reasoning and high learning systems &
Regularisation models &
Symbols are added to objective function during training. &
Can prevent overfitting by handling noisy data. \\
& Knowledge transfer models &
Knowledge Graphs (KGs) are integrated with neural architectures. &
Zero-shot learning and few-shot learning. \\
\bottomrule
\end{tabular}
\end{table}
\section{Neurosymbolic AI (NeSy AI)}

\subsection{NeSy AI Approaches}
\label{subsec3.1}
\begin{figure}[t]
\centering
\includegraphics[width=0.60\textwidth]{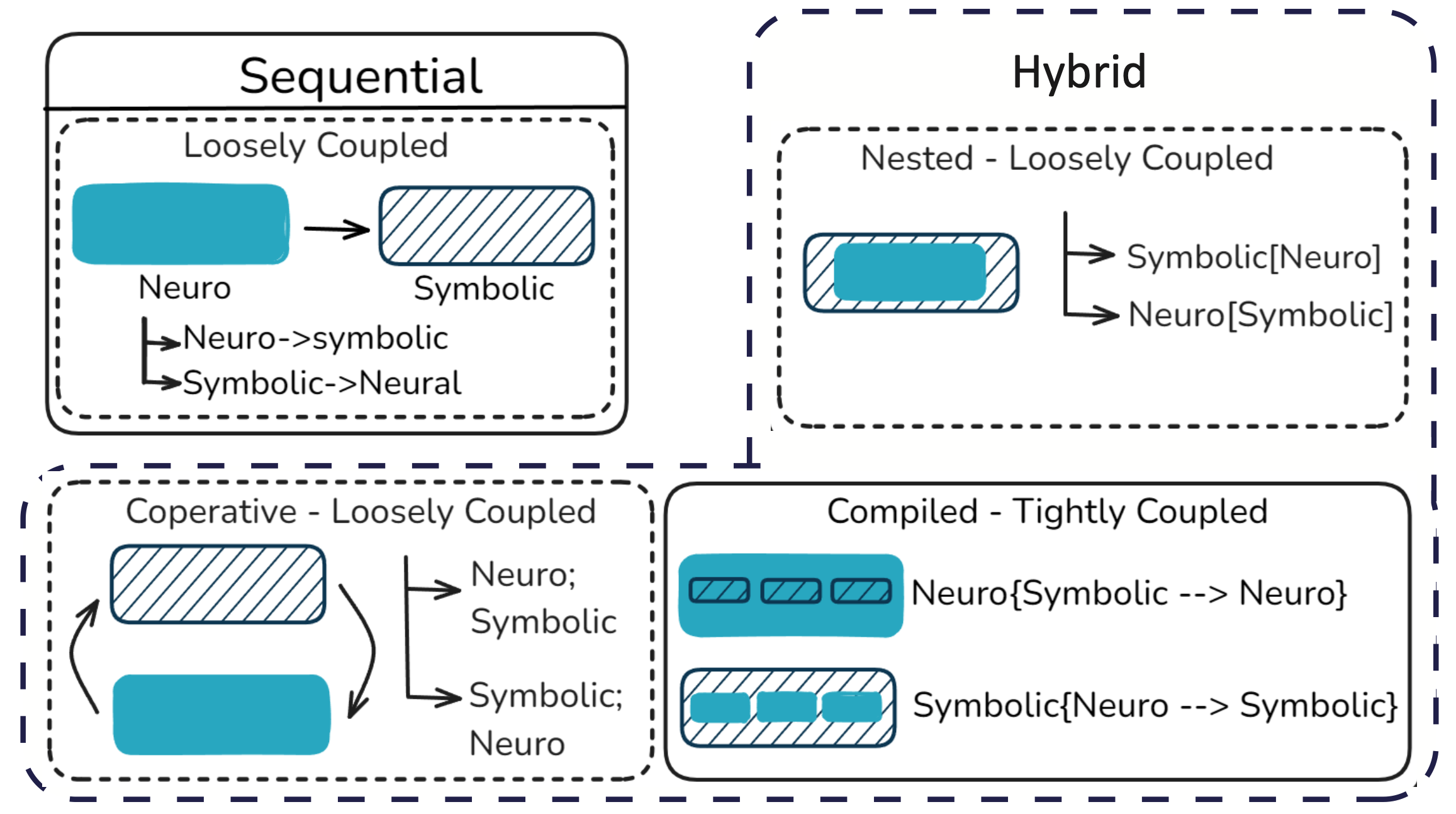}
\caption{Neurosymbolic Approaches depending on the interaction level}\label{fig:coupling}
\end{figure}

NeSy AI integrates neural architecture, such as neural networks, with symbolic solvers. Symbolic solvers refer to KGs (KGs) and various logic systems, including First-order logic (FOL), Fuzzy logic (FL), Propositional logic (PL), and Rule-based systems (RB).
\begin{itemize}
\item FOL is a formal system that represents objects, their properties, and relationships in quantified variables, such as “for all” and “there exists”, to refer to specific elements within a domain. 
\item FL is used for reasoning under uncertainty, utilising truth values that range between 0 and 1, rather than being limited to the binary values of true or false. 
\item PL is a basic logical system which represents statements as true or false. To form more complex expressions, these statements can be combined using logical connectives like “and” “or,” and “not.” 
\item RB systems are AI systems that rely on predefined “if-then” rules to infer conclusions or trigger actions. 
\end{itemize}
NeSy approaches are typically classified based on the interaction between neural networks and symbolic reasoning. These include sequential, nested, cooperative, and compiled models, each offering different ways to integrate learning and reasoning components [19], presented in Figure~\ref{fig:coupling}. Sequential is the standard deep learning approach, where neural and symbolic components perform consecutively in a series. In nested approaches, the symbolic solver performs logical tasks using a deep learning approach. In cooperative approaches, the neural component interacts with the symbolic component to make certain decisions. In compiled approaches, symbolic logic is incorporated into the training set of the neural model. In tightly coupled systems, both symbolic and neural components are integrated tightly to coordinate performance. From the perspectives of reasoning and learning, NeSy systems are classified into two categories: i. High Reasoning and Low Learning, and ii. Low Reasoning and High Learning. In high-reasoning and low learning systems, neural networks make relatively fewer contributions and instead use symbolic representations of problems to make predictions with relatively less assistance from neural models [22]. Whereas low-reasoning and high-learning systems use neural networks to make predictions, and symbolic reasoners are integrated at training phases with relatively low contributions [23]. Different approaches to performing various tasks for high-reasoning, low-learning, and low-reasoning, high-learning systems are mentioned in Table 2.

\subsection{Frameworks for Neurosymbolic AI}
\label{subsec3.2}
A framework is a structured approach that provides essential components and guidelines for developing algorithms [27]. Popular frameworks for NeSy AI include Logical Tensor Networks (LTN) [27], DeepProbLog [22] , Neural Logic Machines (NLM) [29], and Neural Theorem Prover (NTP) [30]. 
LTNs integrate logic with neural networks by representing logical formulas as tensors and applying them within a neural network. This approach enables reasoning and learning over structured data, leading to improved inference.
DeepProbLog enhances the probabilistic logic programming system by incorporating deep neural networks. This framework enables the system to learn probabilistic models directly from data, amalgamating the expressive capabilities of logic programming with the adaptability of deep learning. It caters to tasks that necessitate both reasoning and managing uncertainty.
NLMs combine neural networks with symbolic logic using a differentiable logic engine. This enables the model to perform logical learning and reasoning tasks using tensors activated by logical rules, such as deductive, inductive and abductive reasoning. 
NTP integrates symbolic reasoning with deep learning methodologies to analyse mathematical theorems. It establishes their validity by identifying patterns and generalisations from extensive datasets comprising known theorems and proofs. This neural network-based automated theorem-proving strategy employs Prolog syntax to conduct logical reasoning by embedding logic within neural networks.
These frameworks enable machines to reason and learn by converting rules expressed in first-order logic into a form that neural networks can process. Such frameworks combine learning and reasoning by representing knowledge as tensors and applying logical rules. This helps the AI to generalise from specific cases, improving its reasoning ability. Each framework aims to bridge the gap between traditional logic-based AI and neural networks, allowing systems to handle structured, logical reasoning and more complex, unstructured data.
Following the framework discussion, an architecture analysis of NeSy AI is crucial for examining the strengths of both neural and symbolic components. Table 3 presents the architecture, focusing on the syntactical and semantic details of these frameworks. The syntactic aspect encompasses state propositional, first-order, or relational logic. The semantic approaches highlighted here are fuzzy and probabilistic, employed to manage knowledge representation and reasoning under uncertainty. Fuzzy logic addresses approximation, while probabilistic logic handles uncertainty in knowledge representation. The inference mechanisms facilitate making predictions based on data from various task types they are designed to handle. Table 3 outlines proof-based and model-based inference mechanisms for different knowledge representation, induction, and generative tasks. Proof-based inference utilises formal logic to derive conclusions, whereas model-based inference relies on probabilistic methods. The specific architecture, suitable for multiple paradigms including logic, probabilistic reasoning, and neural approaches, allows researchers to choose frameworks aligned with their research objectives. This can further foster innovation by encouraging researchers to explore novel combinations or adaptations of existing frameworks, thereby opening new avenues for research in the architecture of NeSy AI frameworks.

\subsection{Neurosymbolic AI (NeSy AI) vs Symbolic-integrated LLM}
\label{subsec1}

\begin{figure}[t]
\centering
\includegraphics[width=0.60\textwidth]{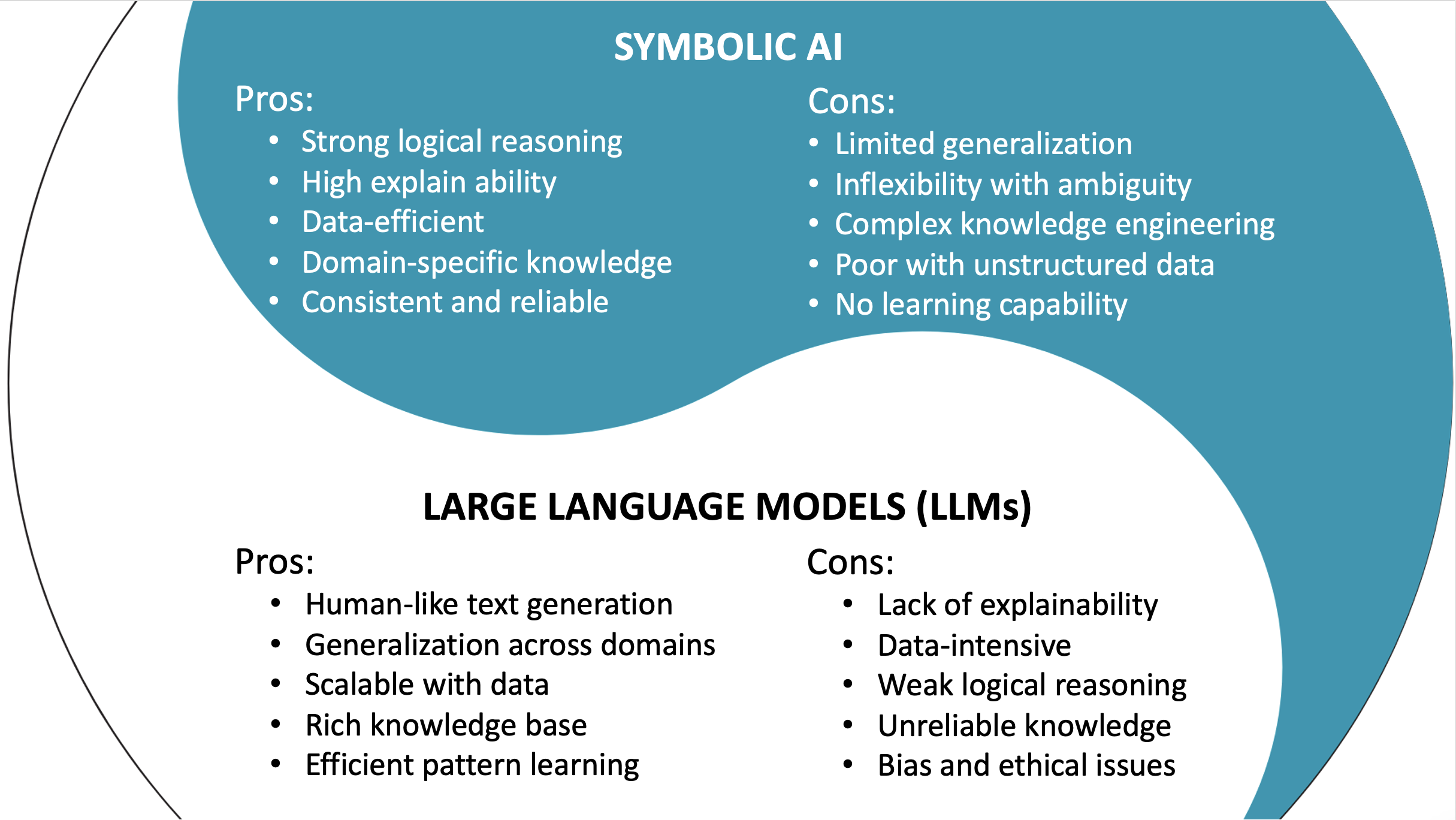}
\caption{Strengths and Weaknesses of Symbolic AI and LLMs}\label{fig:SandW}
\end{figure}

NeSy AI approaches, such as high-reasoning/low-learning systems and low-reasoning/high-learning systems, were designed for traditional neural networks and have proven effective [31]. In such approaches, symbolic and neural components are coupled through joint training or differentiable logic layers. NeSy frameworks like DeepProbLog, Logical Tensor Networks, NLM, and NTP are compatible with standard neural architectures and are potentially more powerful for deep, compositional reasoning [32]. They are mostly conceptual at present, with fewer real-world, large-scale implementations. Considering the strengths and weaknesses of symbolic systems and LLMs as illustrated in Figure~\ref{fig:SandW}, this integration can result in more practical and trustworthy solutions [33]. However, integrating LLMs with NeSy approaches and frameworks to enhance LLM capabilities is an early research stage, with some existing approaches pursuing theoretical perspectives. A notable theory-focused study by A. Sheth et al. presents a theoretical framework for NeSy integration with LLMs to enhance algorithmic capabilities such as reasoning and application-level features such as explainability [34]. Apart from theory-focused approaches, some approaches target application-level LLM-based NeSy integration. However, these approaches mainly focus on improving the NeSy part by utilising LLM instead of LLM enhancement [35]. One strategy that utilises LLMs for automatically extracting features to enhance symbolic parts is outlined in [38], which employs enhancement in LLMs to strengthen the symbolic components rather than incorporating NeSy elements to augment LLMs.

LLMs operate on a transformer-based architecture rather than standard neural networks and do not require retraining of the symbolic component. The pretraining–finetuning paradigm limits direct symbolic coupling at the learning stage. In practice, symbolic integration with LLMs is typically performed by fine-tuning, prompt engineering, or external knowledge injection, rather than the joint training strategies. 
Therefore, NeSy general frameworks are not directly transferable to LLMs, and symbolic-integrated LLMs can be viewed as a specialised branch of NeSy AI. This requires tailored integration strategies beyond conventional NeSy frameworks. This sub-branch further requires unique adaptations. This review targets existing literature for NeSy AI and then explores the integration of symbolic AI with LLMs, considering the need for improved transparency from symbolic AI. For the remainder of this paper, the term NeSy is used in a narrowed sense, specifically to denote the branch concerned with symbolic integration in LLMs.Table 4 presents a comparative analysis of diversed symbolic-integrated LLMs. By examining the performance and limitations of each approach, the study highlights how different integration strategies affect LLMs' capability and scalability. The focus is on evaluating the strengths and weaknesses of each method in terms of reasoning, explainability, flexibility, and computational efficiency. This comparison offers insight into the current state of NeSy integration and identifies areas for improvement to optimise LLMs for more complex, interpretable, and robust AI systems.  

\section{Symbolic-integrated LLMs}
\subsection{Integration Stages}
\label{sec4}

In NeSy, symbolic integration with neural networks is carried out at both the architecture and training levels by embedding logic into the learning process. Meanwhile, integration with LLMs considers symbolic embedding at the application and interface levels. Hence, integration stages differ in LLM-symbolic integration. 
A range of existing surveys explore symbolic integration with neural approaches without giving sufficient weightage to LLM integration. A recent survey paper examines the symbolic integration of neural approaches over the past two decades, drawing on a wide range of sources, including books [42]. It investigates learning, reasoning, scalability, explainability, decision-making, and ethical considerations from a neural perspective without placing significant emphasis on application or tool-level integration, such as LLMs [43]. Another existing systematic literature review examines the application of symbolic knowledge extraction and injection methodologies in neural approaches, with a limited focus on integration stages.  

Integrating LLMs with symbolic approaches, specifically KGs, is a rapidly evolving phase. Other than KGs, symbolic approaches include rules-based systems, logic programming, and theorem provers. KGs and logic integration are prioritised in symbolic logic integration with LLMs due to their scalability, flexibility, and ability to complement LLMs' probabilistic nature while enhancing reasoning and explainability. In contrast, rule-based systems and theorem provers are often rigid, computationally intensive, and less adaptable to unstructured, dynamic data. This section discusses integrating two symbolic approaches: KGs and logic-based systems with LLMs. It concludes by reviewing the state-of-the-art advancements in both techniques and highlighting the challenges associated with each method. 
\begin{table}[htbp]
\scriptsize
\renewcommand{\arraystretch}{1.3} 
\setlength{\tabcolsep}{5pt} 
\caption{Architecture of Popular Frameworks for Neurosymbolic AI}\label{tab:frame}
\begin{tabular}{p{1.8cm}p{1.8cm}p{1.5cm}p{1.8cm}p{1.5cm}p{2cm}p{1.5cm}}
\toprule
\textbf{Frameworks} & \textbf{Architecture} & \textbf{Syntax} & \textbf{Learning} & \textbf{Inference} & \textbf{Tasks} & \textbf{Paradigms} \\
\midrule
Logic Tensor Network (LTN) &
Symbolic and Neural &
First Order Logic &
Parameterized &
Model &
Semi-supervised, Distant supervised &
Logic, Neural \\
Neural Logic Machine (NLM) &
Symbolic &
Relational &
Parameterized and Structured &
Proof &
KG Completion and Knowledge induction &
Logic, Neural \\
Neural Theorem Prover (NTP) &
Symbolic and Neural &
Relational &
Parameterized and Structured &
Proof &
KG Completion and Knowledge induction &
Logic, Neural \\
DeepProbLog &
Symbolic and Neural &
First Order Logic &
Parameterized and Structured &
Model and Proof &
Distantly supervised Knowledge induction &
Logic, Probability, Neural \\
\bottomrule
\end{tabular}
\end{table}

\begin{figure}[t]
\centering
\includegraphics[width=0.60\textwidth]{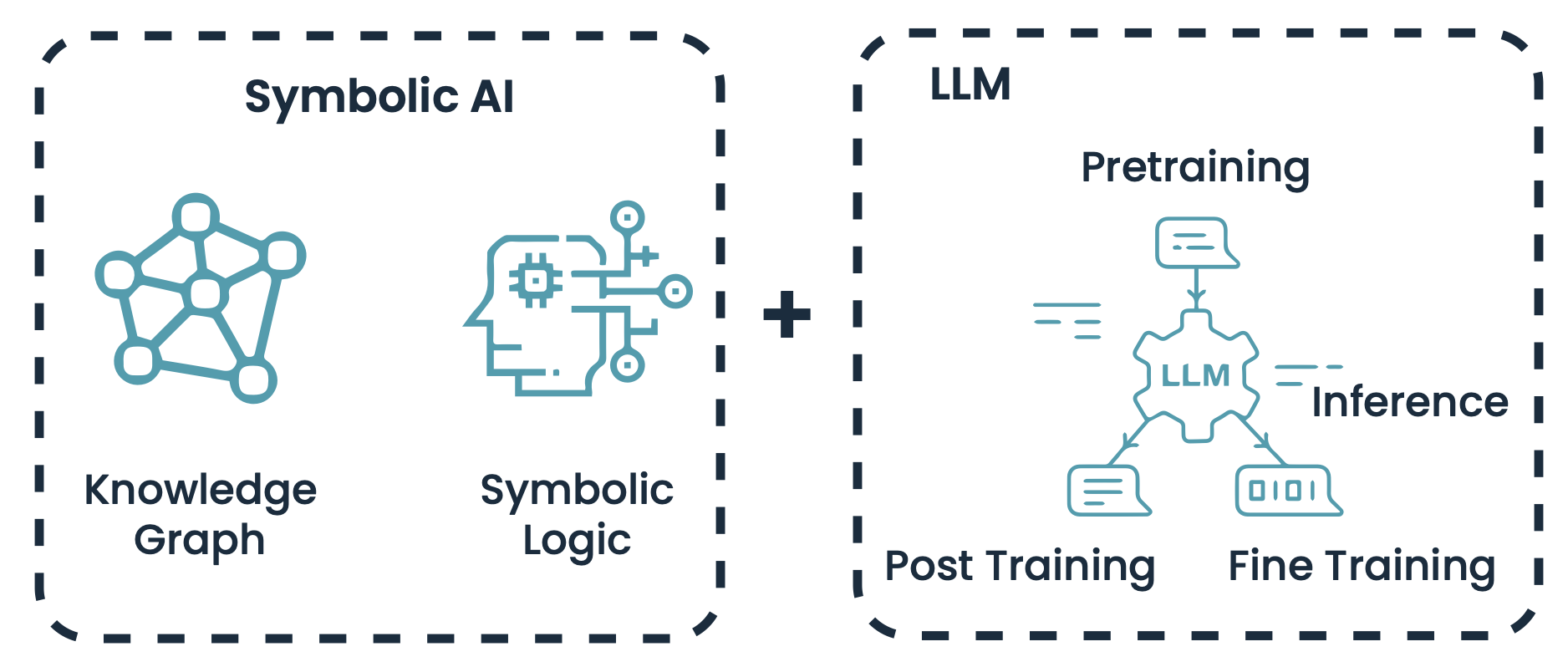}
\caption{Symbolic-integrated LLM}\label{fig:integration}
\end{figure}

Each approach's integration is explored at several stages, including pre-training, inference, and post-training or fine-tuning. This stage-specific integration architecture offers unique benefits and challenges. In the pretraining phase, KG integration enhances LLM learning from augmented textual and structured training data, resulting in better relationship analysis from entities [46]. KG can expand input structures, convert textual inputs into trees, enrich input information using text embeddings, and optimise Word Masks [47] [48].  KG can also be incorporated during the training phase of LLM by modifying different transformer components to improve contextual understanding, including the self-attention mechanism. Different approaches include knowledge-infused self-attention transformers [49], focus fusion attention mechanisms [50], and dual-interleaved attention [51]. 
During the training phase, models incorporate additional adapters [52] and specific encoders for KG [53] or insert additional encoding layers [54] to perform reasoning and enhance accuracy. At the inference phase, KG can be integrated using Retrieval Augmented Generation (RAG) to provide further context and improve factual accuracy [55]. Post-processing integration at the post-training phase is performed to validate content accuracy through concept net [56] or generating knowledge-based prompts [57]. Another complete end-to-end LLM-symbolic planner without expert intervention integrates symbolic AI into the inference stage of LLM [58]. 
In the pre-training stage, logic rules are infused with training data for structured learning patterns to improve understanding of logic and reasoning [59]. This stage includes approaches like embedding logical constructs [60] and logic-guided augmentation of symbolic data into models [61] vocabulary and pre-training corpus. In Logic-guided fine-tuning, the model specialises in logic building without altering the architecture. Symbolic logic can be integrated into the inference phase using external symbolic solvers in the inference pipeline for symbolic reasoning [59].  
Figure~\ref{fig:integration} illustrates the architecture of symbolic-integrated LLMs, showcasing symbolic components like knowledge graphs and logic integrated across multiple stages, including pre-training, post-training, fine-tuning, and inference, highlighting the flexibility of integration.

\subsection{LLMs and Symbolic Coupling }
\label{sec4.1}

LLM integration with symbolic AI, based on the level of interaction between both components, refers to coupling. Coupling is performed to gain maximum strength in learning, reasoning, and explainability from both paradigms. Coupling offers solutions for various LLM challenges but introduces further challenges that must be addressed. These challenges include design complexity and balancing LLMs' flexibility with the rigidity of symbolic approaches [62]. Coupling categories based on interaction level include Decoupled integration and Intertwined integration.
In Decoupled integration, LLM and symbolic components act as autonomous elements that function independently while interacting when necessary. The autonomous structure promotes flexibility and modular design and enhances scalability.  This integration does not overcomplicate the system's design and implementation, but can limit the depth of integration. The independence can lead to limited interaction and suboptimal performance, resulting in constrained contextual understanding [63]. Decoupled integration can be loosely coupled with little interaction or moderately coupled by sharing data or functionality while maintaining independence. In Intertwined integration, LLM and symbolic components are seamlessly and tightly coupled. Such integration can be thoroughly intertwined without any independence or functionally intertwined while maintaining little independence. This makes the system capable of performing complex reasoning by utilising the maximum strength of both systems, resulting in a complex architecture but better overall performance. The interdependence between the systems challenges the adaptability and flexibility for modification. 

In integrating symbolic AI approaches with LLMs, the coupling can occur at various stages, including pre-training, training, inference, and fine-tuning. In the pre-training stage, the symbolic approaches are integrated before training the LLM to influence the learned representations but are not deeply intertwined with the model's core architecture. Symbolic elements are incorporated as part of the model’s training data. Coupling at this stage is usually loosely or moderately coupled. DKPLM is a moderately coupled knowledge-enhanced pre-trained language model that integrates KG at the training stage to inject external knowledge triples from KGs [64]. KEPLER is a knowledge embedding and pre-trained language representation model in which KG is integrated into the training objective at the pretraining stage with moderate-level coupling[65]. Similarly, deterministic LLM pre-trains language models with deterministic factual knowledge by integrating KG into the training objective [66]. Whereas Symbol-LLM represents a tightly coupled approach to integrating symbolic knowledge into the injection and infusion stage of LLM for a deep understanding of symbolic knowledge in LLM parameters [67]. Other approaches incorporating KG into the input of the language model at the pre-training stage, where the coupling is moderate or loose, include K-Bert[48], CoLAKE[68] and ERNIE3.0 [69].  In inference, symbolic elements are incorporated as training objectives to adjust weights and guide the model’s learning process. The level of dependency and interaction can result in moderate to tight coupling. LLM-Modulo Framework is an inference-level approach with a tight coupling of symbolic and LLM components leveraged to perform reasoning and model-based planning[70]. Another tightly coupled KG-integrated LLM approach for knowledge-based reasoning and back traceability, where inference is performed without explicit re-training, is explored in [71]. 

\begin{sidewaystable}
\scriptsize
\caption{Comparative Analysis of Symbolic-integrated LLM Approaches}\label{tab:landscape-example}
\renewcommand{\arraystretch}{1.2} 
\begin{tabular*}{\textheight}{@{\extracolsep\fill} p{0.5cm} p{1.8cm} p{1.5cm} p{1.8cm} p{2cm} p{2.5cm} p{4cm} }
\toprule
\textbf{Study} & 
\textbf{Task } & 
\textbf{Neural Part} & 
\textbf{Symbolic Part} & 
\textbf{Dataset} & 
\textbf{Performance} & 
\textbf{Limitation} \\
\midrule

[33]	&	Logical reasoning in natural language 	&	StarCoder+, GPT-3.5, GPT-4	& FOL and deductive inferring	&	FOLIO and ProofWriter	&	 26\% higher accuracy  with GPT-4 than CoT on ProofWriter (FOLIO)	&	Only one aspect of logical reasoning is focused, Scalability, computational costs, Better reasoning and logic techniques can be used	\\
{[34]}	&	Mathematical Reasoning and symbolic interpreter	&	CODEX, COT (Minerva, PaLM)	&	Arithmetic and symbolic reasoning &	  GSM8K, SVAMP, ASDIV, MAWPS,  BIG-Bench	&	GSM8K outperforms (COT by absolute 15\% top- 1 accuracy.)	&		\\
{[35]}	&	Reasoning in Machine Reading Comprehension	&	GPT3.5 Turbo,  LLaMa 7B,  Alpaca 7B	&	Numerical reasoning 	&	Manually created dataset	&	Improved performance for all LLMs	&	Extend the task to allow for nested reasoning and learning commonsense knowledge to solve complex questions beyond numerical reasoning	\\
{[36]}	&	Translating language into code to solve programming tasks 	&	Codex gpt-3.5-turbo-0301, and GPT-4	&	Symbolic compression module	&	REGEX, CLEVR, LOGO	&	LOGO 32.13\% , CLEVR 88.67\% , REGEX   76\% 	&	Search does not condition on Input output examples, instead it focuses on patterns in description solution pairs	\\
{[37]}	&	Storytelling	&	OpenAI’s Code-LLM(GPT3), Codex 	&	Reason & Prompt		&	ROC-AUC score CoRRPUS (specific functions) 79\%	&	CoRRPUS produced code is not guaranteed to run and completely accurate in reasoning	\\
{[38]}	&	Automated agents and Symbolic Reasoning in Games	&	OpenAI LLMs	&	Rule based reasoning for math, map reading, common sense, sorting	&		&	88\%	&	More explained prompts can enhance LLM agent actions, symbolic modules can be added to handle complex scenarios	\\
{[39]}	&	Math word problems solver	&	Codex	&	External symbolic solver to solve equations	&	GSM8K math problems, new dataset ALGEBRA	&	Outperforms (PAL) by 20\% (on ALGEBRA)	&	Declarative prompts with SymPy, a Python library for symbolic computation	\\
{[40]}	&	General	&	BART	&	PL, Inference as forward and backward chaining	&	SimpleLogic dataset (PL)	&	Label-Priority, Rule-Priority, Balanced Rule priority	&	Bottom-up and top-down approaches are not fully comparable. 	\\
{[41]}	&	Physics	&	Open AI’s codex model	&	linguistic reasoning benchmark	&	Manually created dataset	&	R2 = 0.759, p < 0.00 Correlations with human judgements	&	Integration with pragmatic interpretation, inference, and perception, to construct better representations from perceptual inputs can enhance results	\\

\botrule
\end{tabular*}
\end{sidewaystable}
Yet another tightly integrated approach in which LLM acts as the semantic parser for processing tasks on natural and formal language instructions is discussed in [33]. LAAs are another tightly coupled symbolic integration in LLM-empowered agents deployed to enhance inference reasoning [72].
In the fine-tuning stage, symbolic components influence the task-specific requirements without altering the model’s core architecture, resulting in moderate coupling. The symbolic component balances the influence of symbolic knowledge with the model’s learned patterns. The GLaM is a KG-integrated LLM approach which uses graph encoding to fine-tune LLM[73].  ToM-LM is another fine-tuned LLM on natural language and symbolic representation to generate the symbolic formulation [74]. ChatKBQA is a KG integration-based approach that uses PEFT techniques to fine-tune LLMs to generate logical forms [75]. RoG is another moderate to tightly coupled fine-tuning approach that integrates LLMs with KGs to enable reasoning [76]. Integration at the inference stage is usually loosely coupled. At this stage, the output generated from LLM is validated and altered by the symbolic component depending on the application without modifying the model’s layer. The most widely used approaches in the literature for integrating symbolic AI are usually performed by RAG or prompt-based approaches[77]. LoT is a prompt-based logic integration approach which uses propositional logic to enhance reasoning [78]. Prompt-based inference-level KG integration approaches include CoK [79], ChatRule[80] and Mindmap [81]. TC–RAG is a RAG-based framework that integrates symbolic logic with LLMs to enhance the RAG process[82]. Other KG-based RAG approaches include KGLM [83] and REALM[84]. 

\subsection{Application level and Algorithm level Symbolic-integrated LLM.}
\label{sec4.2}

This section explores the integration of LLMs with symbolic AI at the algorithmic or application level. Each approach offers unique opportunities and challenges. From application-level and algorithm-level integration perspectives, the reviewed works broadly discuss NeSy approaches that target neural architecture with symbolic data without integrating language models[85]. These approaches include KB-ANN [86], Logic Tensor Networks [28], and Logic Neural Networks [87]. However, algorithm-level and application-level categorisation targeting LLM with symbolic approaches is less commonly explored. These less prevalent approaches focus more on KG than rules-based and logic-based systems. However, several established techniques encompass both perspectives effectively, including embedding KGs with LLM using vector embeddings, adapters, or masking-based methods, which are broadly discussed in the literature [88], [89], [90]. Table 5 presents a summarised application-level and algorithm-level perspective.

\begin{table}[htbp]
\scriptsize
\caption{Application-level and Algorithm-level Perspective}\label{tab:integration-comparison}
\renewcommand{\arraystretch}{1.3} 
\begin{tabular*}{\textwidth}{@{\extracolsep\fill} p{2cm} p{5cm} p{5.5cm}}
\toprule
\textbf{Criteria} & \textbf{Application-Level Integration} & \textbf{Algorithm-Level Integration} \\
\midrule
	Integration Level	&	Higher-level integration, typically at the system or workflow level	&	Lower-level integration, directly within the LLM's architecture	\\
	Primary Focus	&	Enhancing LLM capabilities through external symbolic components	&	Incorporating symbolic reasoning into the core LLM functioning	\\
	Implementation Complexity	&	Generally lower, as it involves combining existing tools	&	Higher, requires modifications to the LLM's internal structure	\\
	Scalability	&	Often more scalable, as components can be added or removed	&	 Less scalable due to increased computational requirements	\\
	Explainability	&	Typically higher, as symbolic components provide clear reasoning steps	&	 Challenging, as integration occurs within the neural network	\\
	Performance Impact	&	 Introduce latency due to external processing	&	 Faster execution, but may require more training time	\\
	Flexibility	&	More flexible, easier to adapt to different tasks or domains	&	Less flexible, changes require retraining the entire model	\\
	Coupling with LLM	&	Shallow, external system-level queries	&	Deep, integrated into training/fine-tuning	\\
	Use Cases	&	Task-specific enhancements, domain-specific knowledge integration	&	Improving general reasoning capabilities, enhancing model interpretability	\\
	Performance	&	Moderate, task-dependent	&	High, enhanced reasoning capabilities	\\
	Complexity	&	Low	&	High	\\
	Resource Requirements	&	Low	&	High	\\
	Challenges	&	Ensuring seamless integration, managing increased system complexity	&	Balancing symbolic and neural components, preserving LLM's generalization ability	\\
\bottomrule
\end{tabular*}
\end{table}

\subsubsection{Application-level Integration and Features}
\label{sec4.2.1}

Application-level integration refers to integrating symbolic approaches and knowledge-based systems to enhance AI capabilities in domain-specific and task-specific real-world applications. This refers to the practical implications and functionalities that perform well algorithmically and consider the end user’s requirements while producing output. This includes compressing KGs, symbolic rules, or logic for integration at a particular phase to enhance LLM capabilities.  Rules can be deduced from expert systems or theorem provers. Logic can be first-order logic, predicate logic, or propositional logic. Logic-based LLM integration at the application level includes approaches like SymbolicAI [33], FOLIO [91], LINC[86], [87], and Logic LM [59].  
These features include multi-disciplinary adaptation to domains using domain-specific knowledge such as ontologies, KGs and rule-based systems. The application level emphasises user-level features, including explainability, interpretability, safety, and trust[34]. At the application level, user explainability is a crucial feature for building the confidence of stakeholders and supporting scalability for diverse use cases. Domain-specific constraints, applicability, and continual learning to ensure adaptability in diverse domains are essential features at the application level of NeSy. Application-level integration emphasises logical consistency and robustness within LLM responses by minimising errors that LLMs can generate when working independently. Real-time knowledge upgradation in LLMs using dynamic KGs or KBs is another notable feature of application-level integration. These features make application-level symbolic integration a powerful approach for enhancing LLMs' accuracy, interpretability, and reliability in real-world settings.

\subsubsection{Application-level Integration and Features}
\label{sec4.2.2}
Algorithm-level integration refers to LLM processing pipelines' computational and technical aspects [92]. Technical aspects focus on tightly embedded symbolic approaches into LLMs’ fundamental architecture or training process. Algorithm-level features are more tightly integrated into model functioning[93]. These features are learnt and optimised during training to interact seamlessly. The seamless incorporation of features into LLM’s architecture makes modification challenging. Modification is possible only by altering the language model's or transformer's core architecture using specific symbolic algorithms. The LLM-enhanced embedding approach in [94] is an algorithm-level integration. KiL is another algorithm-level integration containing context-adaptive algorithms to infuse knowledge into neural architectures [95]. 
These features include enhancing abstraction, analogy, reasoning and planning[34]. Abstraction is the ability to summarise the information by generalising it for high-level concepts. Algorithm-level integration can effectively organise knowledge hierarchically based on abstract ideas rather than concrete details. Abstraction is crucial in high-risk sectors; for instance, in medical diagnostics, an LLM could abstract specific symptoms into broader categories (e.g., “respiratory issues”), facilitating efficient diagnosis across related conditions. Analogical reasoning enables LLMs to identify shared structures with similarities between different scenarios, using symbolic integration to map similarities across knowledge domains. This supports creative problem-solving and generalisation. Planning allows an LLM to anticipate steps needed to reach a goal, using symbolic integration to optimise decisions across multiple stages. Algorithm-level integration with symbolic components will enable LLMs to leverage planning algorithms, such as Markov decision processes or symbolic task planners, to simulate outcomes and select optimal actions.

\subsubsection{Comparative analysis of algorithm-level and application-level integration}
\label{sec4.2.3}

In algorithm-level integration, features are embedded into the architecture, resulting in in-depth implementation. This contradicts application-level features such as explainability, which are usually added as additional processes and result in shallow implementation. The operational focus at the application level is based on user interaction, which is contradictory to internal implementations at the algorithm level. Adaptability applies to the application level, whereas generalisation across various domains applies to the algorithm level. The algorithm level is less flexible, with relatively high complexity in design and implementation, but. It is more efficient and cohesive. 
In contrast, the application level is more flexible in terms of adding new modules. The algorithm level offers potentially better scalability due to a unified architecture; however, the application level may face challenges in scaling due to its modular nature. Both integrations present a spectrum of approaches, simultaneously offering significant strengths and challenges. 

This section provides an overview of the integration strategies employed in combining symbolic AI and LLMs, emphasising the stages of integration, coupling mechanisms, and the application-versus algorithm-level approaches. The analysis of existing studies has identified key insights into their methodologies and limitations. A selection of studies demonstrating these approaches has been analysed, with their key characteristics and integration mechanisms presented in Table 6. These findings offer a brief overview of existing approaches while highlighting opportunities for future research. For instance, most of the approaches target integration at the inference stage with moderate or loose coupling and tend towards favouring application-level integration. This analysis indicates the need for further research focused on the pretraining, fine-tuning, and training stages. Additionally, the absence of tightly coupled approaches at the algorithmic level points to another potential direction for enhancing symbolic-LLM integration. 
\begin{table}[htbp]
\scriptsize
\caption{Symbolic-integrated LLM approaches illustrating coupling mechanisms, integration stages and application vs algorithmic level mechanisms}\label{tab:coupling}
\renewcommand{\arraystretch}{1.2} 
\begin{tabular*}{\textwidth}{@{\extracolsep\fill} p{0.7cm} p{3cm} p{3cm} p{1.5cm} p{1.6cm} p{1.5cm}}
\toprule
\textbf{Study} & \textbf{Description} & \textbf{Symbolic Part} & \textbf{Coupling} & \textbf{Integration Stage} & \textbf{Integration Level} \\
\midrule
    {[33]}	&	LLM, a semantic parser to perform deductive inference by FOL expressions	&	FOL to perform Logical reasoning, and deductive inferring	&	Moderately coupled	&	Inference	&	Application 	\\
	{[34]}	&	Mathematical Reasoning and symbolic interpreter	&	Arithmetic and symbolic reasoning tasks	&	Loosely Coupled	&	Inference 	&	Application 	\\
	{[35]}	&	Symbolic AI employed to learn rules to reconstruct answer	&	Numerical reasoning for machine reading comprehension	&	Moderately coupled	&	Inference	&	Application 	\\
	{[36]}	&	Neurosymbolic framework for Library Induction from Language Observations 	&	A symbolic compression module, Probabilistic Context Free Grammar	&	Moderately coupled	&	Inference	&	Algorithm	\\
	{[37]}	&	Information extraction by 1-shot prompting 	&	Reasoning by using Prompting	&	Decoupled	&	Inference	&	Application 	\\
    {[38]}	&	Prompts for common sense tasks, Sorting, math, map reading,	&	Symbolic Reasoners	&	Decoupled	&	Inference 	&	Application 	\\
	{[39]}	&	LLM Integration with symbolic solvers to solve math word problem	&	Symbolic Solvers	&	Decoupled	&	Inference 	&	Application 	\\
	{[40]}	&	To improve inference with BART, supporting logical reasoning.	&	Propositional logic, Inference as forward and backward chaining	&	Decoupled	&	Inference	&	Application 	\\
	{[41]}	&	Fuzzing (automated bug-finding techniques) by LLMs	&	TitanFuzz	&	Decoupled	&	Inference	&	Application 	\\
	{[94]}	&	 LLM fine-tuning	&	Ontology-based	&	Moderately coupled	&	Fine-tuning 	&	Algorithm 	\\
	{[97]}	&	Integration of LLMs and Satisfiability modulo theory 	&	Deductive reasoning engine	&	Moderately coupled 	&	Inference 	&	Algorithm	\\
\bottomrule
\end{tabular*}
\end{table}

\section{Architectural paradigms}
\label{sec5}

The integration of symbolic artificial intelligence with LLMs is classified into various architectural paradigms. The objective is to harness the strengths of both methodologies while alleviating their respective limitations. This section examines three primary architectural paradigms: the LLM-to-Symbolic pipeline, the Symbolic-to-LLM pipeline, and Hybrid Models.

\begin{figure}[t]
\centering
\includegraphics[width=0.85\textwidth]{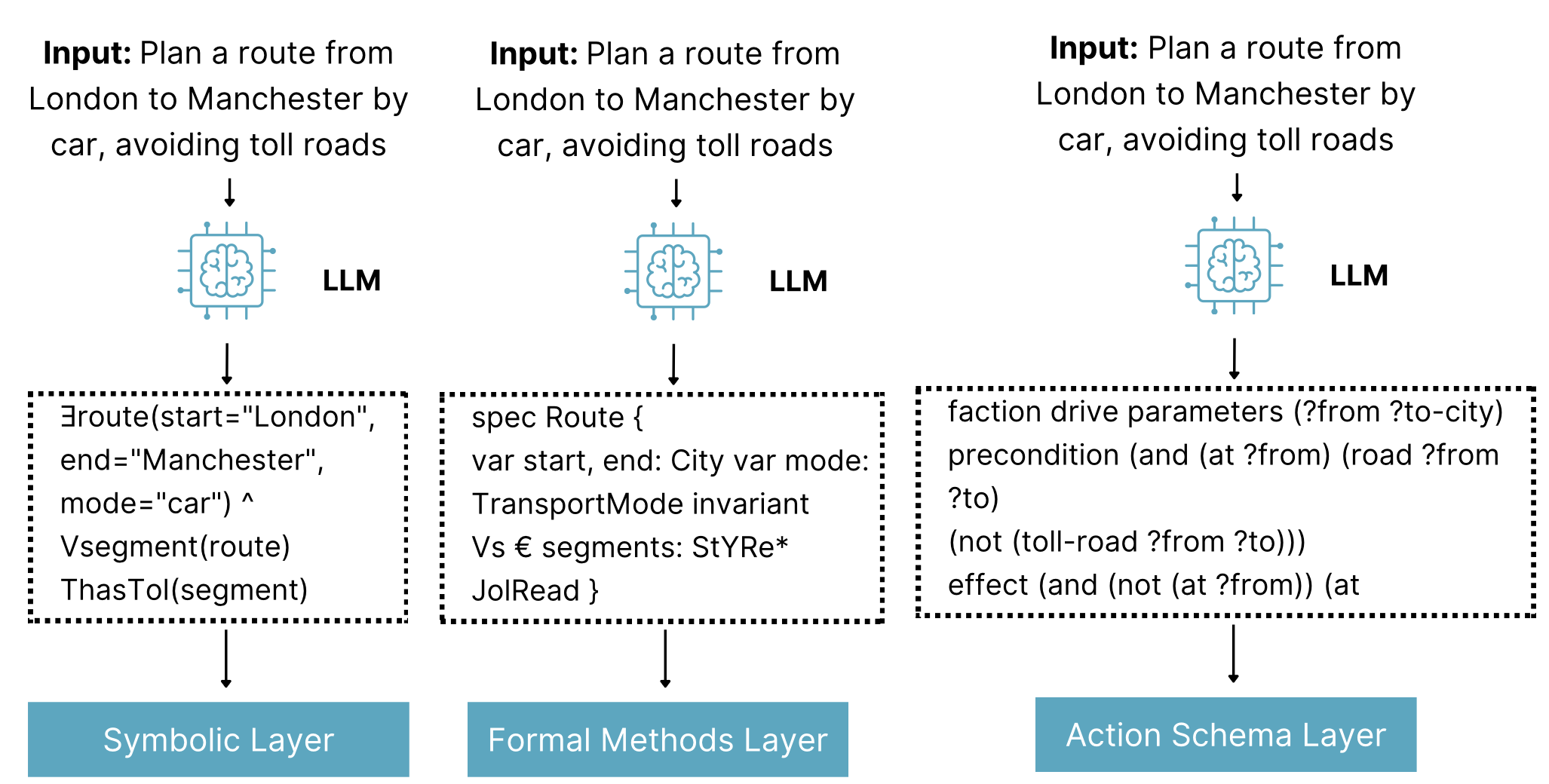}
\caption{LLM to Symbolic Pipeline with three subcategories: a) LLM to symbolic translation, b) LLM to formal Methods Translation, c) LLM to Action Schema Generation}\label{fig6: LLMTo}
\end{figure}

\subsection{LLM to Symbolic Pipeline}
\label{sec5.1}

The LLM-to-symbolic pipeline employs LLMs to create structured symbolic representations. Symbolic engines then process these representations or knowledge structures. Various methods for converting natural language to symbolic form using LLMs have been explored in the literature to improve interpretability and reasoning. The Chain of Thought (CoT) framework is a well-known example of the LLM-to-symbolic pipeline. In CoT, the LLM produces a sequence of reasoning steps in symbolic form to facilitate step-by-step reasoning. LLMs’ capability to generate symbolic representations is explored in the next section. Figure~\ref{fig6: LLMTo} presents an illustration that exemplifies the pipeline from LLM to symbolic representations and their respective subcategories.

\subsubsection{LLMs generating symbolic representations}
\label{sec5.1.1}

 LLMs are deployed to create structured symbolic representations from human-understandable language to enhance the interpretability of complex concepts[98]. When incorporated with LLMs, symbolic representations effectively generate accurate, logical inferences by addressing the challenges in reasoning tasks[99]. These symbolic representations are processed either by integrated symbolic solvers or logic engines, prompt engineering for symbolic structures, fine-tuned LLMs on symbolic structures, schema and ontology-guided LLMs, Program of Thought, incorporating syntactic and semantic parsers, iterative refinement based on feedback from rule-based or symbolic systems or by injecting symbolic knowledge into training stages of LLMs. LINC is a NeSy which uses logical provers for formal logical reasoning by integrating symbolic frameworks[36]. Despite its adequate accuracy and performance, it often suffers from a syntactic and semantic mismatch between logical expression generated by the model and what the prover can process, which needs further solutions. SymbCoT uses logic rules and a chain of thoughts to translate natural language into symbolic representations to enhance reasoning abilities[100]. This approach suffers from performance bottlenecks due to sequential solver operations, which demand parallel processing techniques to reduce bottlenecks. 

\subsubsection{Natural language to formal language translation }
\label{sec5.1.2}

LLMs can automatically generate formal language specifications and proofs by translating word problems. This approach can improve challenges related to interpretability, explainability, and reasoning. nl2postcond is an approach that uses LLMs to convert language specifications into symbolic-based formal methods [101]. Formal-LLM is another approach that translates natural language into formal language when designing an LLM-based agent’s plan [102]. Another strategy for using LLMs to translate mathematical problems into formal verification properties is proposed in [87]. The proposed approach employs self-prompting to perform symbolic reasoning to align with the numeric answer, thus generating an interpretable and verifiable response. Various techniques are proposed to assess LLMS' capabilities in translating formal language syntax. However, these assessment metrics are considered inadequate for determining the safety of formal translation outputs, as demonstrated by the extensive experiments in [103], highlighting a significant research gap in the literature. Furthermore, these methods have additional limitations, such as relying on static models and formalising relatively minor theories. To address this limitation, transformer memorisation [88] has been introduced, but it remains insufficient, underscoring the need for further investigation.

\subsubsection{Action schema generation}
\label{sec5.1.3}

This pipeline generates an action schema from language text to design executable plans[97]. Numerous approaches used for LLM-based action schema generation include few-shot prompting[104], corrective re-prompting[105], fine-tuning [106], agent planning [102], hybrid LLM-symbolic methods [102] integrating predefined logic frameworks, schema libraries[58], and automated validation modules. These action schemas generated by LLMs usually adopt human experts in the system, but the challenge is that human interpretations sometimes may need to align with actual user intent[58]. To address this, the LLM-Symbolic Planning pipeline is proposed without deploying expert humans [58]. This pipeline designs a library of diverse action schema candidates, capturing multiple interpretations of natural language descriptions with semantic validation and ranking modules. However, such approaches struggle because of a need for new evaluation metrics suited to dynamically generated action schema models[58]. The literature widely discusses LLM-generated symbolic planners [107], considering their capability to create intelligent planning systems. However, another approach suggests that LLMs cannot plan and only assist in bidirectional modulo frameworks[70]. This dilemma underscores the need for further research and experimentation to validate the hypothesis and clarify its scope of applicability.

\subsection{Symbolic to LLM Pipeline}
\label{sec5.2}

\begin{figure}[t]
\centering
\includegraphics[width=0.95\textwidth]{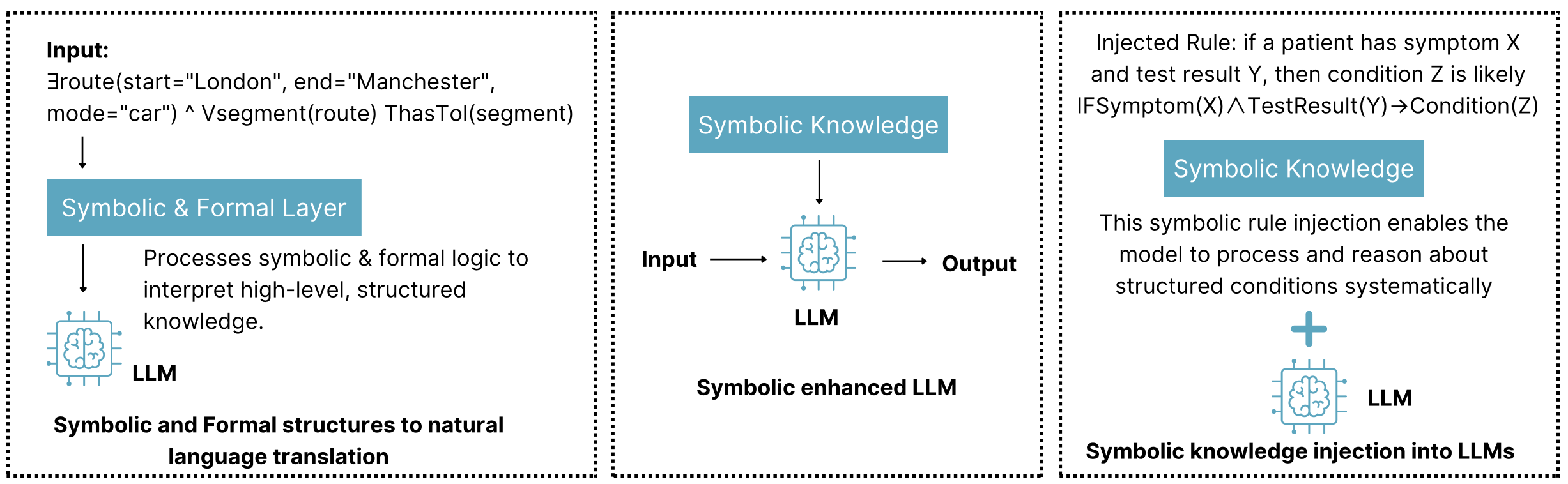}
\caption{Symbolic to LLM Pipeline with three subcategories: a) Symbolic and Formal Structures to Natural Language Translation, b) Symbolic Enhanced LLM c) Symbolic Knowledge injection into LLMs}\label{fig:SymTo}
\end{figure}

The symbolic-to-LLM pipeline integrates external symbolic knowledge into LLMs, encompassing logic rules, KGs, and ontologies. This pipeline effectively empowers LLMs to produce interpretable outputs informed by structured knowledge. Numerous methodologies are explored in the literature to improve interpretability and accuracy within LLMs. A prominent example of a symbolic LLM approach is the Graph RAG, in which relevant information from the KG is extracted and subsequently conveyed to the LLM to facilitate inference generation. Figure~\ref{fig:SymTo} presents an illustration that exemplifies the pipeline from Symbolic to LLM and its respective subcategories.

\subsubsection{Symbolic knowledge injection into LLMs}
\label{sec5.2.1}

Symbolic knowledge incorporation into LLMs can enhance their reasoning capability and explainability.  This knowledge injection can be performed at any stage, including pre-training, fine-tuning, or the inference phase.  Several techniques are proposed to inject symbolic information at various stages of LLMs, including pretraining, fine-tuning, and inference. An approach incorporating structured KGs into LLMs at the fine-tuning stage is discussed in [108], and Kagnet performs knowledge infusion into the fine-tuning stages[56]. Prompting LLMs with injected knowledge at the pretraining stage is discussed in [109]. Injecting symbolic knowledge in training phases can cause a loss of generalisation ability in LLM[67]. A two-stage tuning framework is proposed for injecting symbolic knowledge without catastrophic forgetting[67]. Other knowledge injection approaches include encoding symbolic rules into input prompts, embedding logic and KGs into LLMs, ontology-based injection and prompting, constraint-based fine-tuning, Logic-Informed Fine-Tuning with Symbolic Rules, Schema and Template-Based KG Injection, and hybrid NeSy approaches for validation or refinement. These approaches improve structured reasoning capabilities in LLM. However, challenges remain, including consistency in symbolic injections with the LLM’s probabilistic nature, complexity overhead and scalability. Further research into flexible hybrid architectures could help address these limitations.  

\subsubsection{Symbolic and Formal structures for natural language translation}
\label{sec5.2.2}

Symbolic translation into language using LLMs is an evolving area of research aiming to make complex logical structures more accessible and interpretable. This process can significantly enhance the capabilities of LLMs by providing them with logically structured inputs derived from symbolic AI systems. Various approaches have been explored for translation, including symbolic input parsing by analysing components, relationships and logical connectives, predefined templates for symbolic structure conversion to natural language, and domain-specific vocabulary-based context-aware translation. Various approaches use translated symbolic content to enhance LLM outputs by providing a verifiable reference for additional context. These approaches include supplying logical explanations for complex reasoning steps and improving fact-checking processes in feedback mechanisms and reinforcement approaches, thus serving as a reliable verification layer to support LLM. S2L is an approach which converts symbol-related problems into language-based representations by prompting LLMs or leveraging external tools[109]. Language-based representations are integrated into the original problem, providing in-depth contextual information to assist LLM in enhancing interpretability. Symb-XAI is another approach which converts symbolic logic to natural language to express different logical relationships between input features [110]. Another code-based symbolic translator using Reinforcement Learning with Feedback is discussed in [111].

\subsubsection{Symbolic-Enhanced LLM}
\label{sec5.2.3}

Symbolic-enhanced LLM is another approach explored in the symbolic-to-LLM pipeline for enhancing LLM outputs by leveraging external symbolic systems. The symbolic solvers can be logic engines or KGs to supply structured, factual, or logically consistent information that the LLM doesn’t inherently contain. In the KG-Enhanced LLM Reasoning framework, the KG provides curated, relational data that the LLM does not inherently possess, allowing the LLM to develop more accurate, contextually relevant, or logically sound outputs. In KG-enhanced LLMs, KG retrieves relevant information, which is then used to augment LLM prompts. A KELP framework is proposed to extract useful information from KG and send it to LLM to address the hallucination[112], contrary to approaches where LLMs construct KGs and extract information from KGs[113]. The symbolic system leverages symbolic solvers to perform logical reasoning on pre-defined logical statements and supplies this processed information to the LLM. Apart from common challenges, including data sparsity, coverage, and scalability in annotated logic corpora, this area demands further research to deal with less structured or ambiguous information, such as open-ended queries.

\subsection{Hybrid Models}
\label{sec5.3}

This paradigm covers bidirectional iterative approaches with integrated architectures combining neural and symbolic components in either sequence. This further includes LLM-integrated NeSy architecture and modular systems. The aim is to iteratively leverage the strengths of each system to compensate for the weaknesses of the other. Bidirectional approaches facilitate continuous exchange between LLM and symbolic systems to ensure enhanced explainability. [104] proposes a framework for dynamic collaboration between LLM and symbolic systems to tackle problem-solving and hybrid thinking. Another approach leverages an iterative hybrid architecture to perform continuous reasoning until a successful answer is obtained[114]. Despite better interpretability, such systems need help with the complexity of the interaction pipeline between the components. NeSy combines neural networks' learning capabilities with symbolic systems' reasoning abilities. ProRef is a NeSy approach that applies various prototypes to improve Logical Reasoning using LLMs[115]. Modular systems offer flexible integration, combining plug-and-play components in multiple ways depending on the task. LLM-Modulo Framework is proposed to combine the strengths of LLMs with model-based verifiers for more flexible problem and solution specifications [70]. Hybrid architecture needs more literature targeting seamless integration to address inconsistencies and conflicts in reasoning. The computational complexity associated with integrating symbolic reasoning and large-scale neural networks can lead to heightened latency and resource demands, thereby complicating the efficient implementation of real-time applications. Hybrid models promise tighter integration but can be challenging to design and train effectively. Modular systems offer flexibility but require careful design to ensure effective communication between components.

\section{Benchmarks}
\label{sec6}

\begin{figure}[t]
\centering
\includegraphics[width=0.50\textwidth]{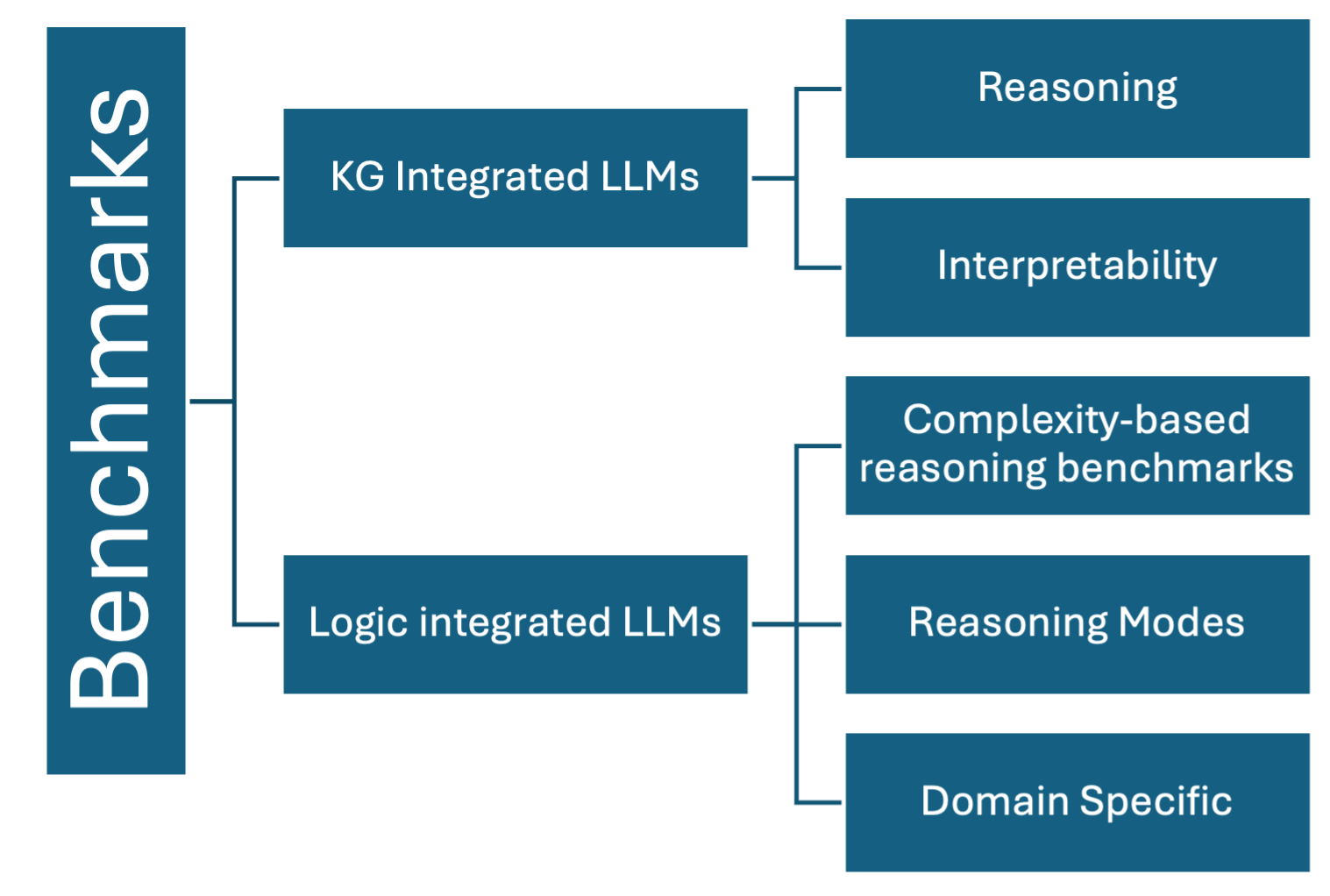}
\caption{Conceptual View of Benchmark Section}\label{fig:benchmarks}
\end{figure}

Benchmarks in symbolic-integrated LLMs function as standardised evaluations for assessing reasoning, explainability, and task performance. This section examines benchmarks related to knowledge graphs (KGs) and symbolic logic. The conceptual framework of the benchmark section is depicted in Figure~\ref{fig:benchmarks}

\subsection{Benchmarks for KGs Integrated LLMs }
\label{sec6.1}

KGs (KGs) offer a structured representation of knowledge, making them particularly valuable for enhancing interpretability and reasoning in LLMs compared to addressing other challenges these models face. This section explores benchmark datasets in existing studies, emphasising reasoning and interpretability using KGs in LLMs. Table 7 presents benchmarks and evaluation metrics used in existing studies integrating KG with LLMs to perform reasoning and interpretability. 

\subsubsection{Reasoning}
\label{sec6.1.1}

Integrating KGs with LLMs has created a necessity to develop benchmarks for evaluating the reasoning abilities of hybrid systems. The Chain-of-Knowledge (CoK) framework assesses and enhances knowledge reasoning in LLMs by integrating KGs [116]. This framework introduces the KnowReason dataset using rule mining and transforming KG triplets into sentences. To assess the step-by-step reasoning and error identification in reasoning steps by analysing assumptions and logic in KG-integrated LLMs, the Meta-Reasoning Benchmark (MR-Ben) introduces human-curated questions across diverse domains[117]. GLUE [118] and SuperGLUE[119] are widely recognised benchmarks for KG-integrated approaches to address reasoning [120]. These benchmarks help evaluate the effectiveness of KG integration, whether used in the embedding phase or to enhance interpretability at various stages [121]. Benchmarks used in other approaches integrating KG for reasoning in LLMs include ATOMIC [122], XNLI [123], ReClor [124], HellaSwag [125], ComplexWebQuestions [126], MetaQA [127], GrailQA [128].  JointLK in [129] integrates KG to perform interpretable reasoning using LM and GNN through a bidirectional attention module and evaluated on CommonsenseQA [130]. GreaseLM [131] uses KG to reason over situational constraints and structured knowledge using three benchmarks: CommonsenseQA [130], OpenbookQA [132], and MedQA-USMLE [133]. A KG-based approach discussed in [134] improves reasoning in LLM on Natural Language Inference (NLI) problems for the science domain, evaluated using the SciTail science questions dataset [135]. KagNet [136] is another approach using KG for commonsense reasoning on the CommonsenseQA benchmark[130]. GAIN [137] uses double graphs with a path reasoning mechanism to provide better reasoning by interpreting relations between entities using the DocRED benchmark [138].  These datasets emulate real-world situations; however, reasoning datasets suffer while analysing assumptions and scrutinising intermediate reasoning steps[117]. Ensuring consistency across reasoning paradigms and generalisation abilities also presents significant challenges [117]. Further efforts are needed to develop more advanced datasets emphasising cross-domain knowledge application, dynamic knowledge updating, and multimodal reasoning[116].

\subsubsection{Interpretability}
\label{sec6.1.2}

The interpretability benchmarks for KG-integrated LLMs focus on assessing the explainability of these hybrid systems. These benchmarks assess the model’s reasoning process by tracing the paths to justify decisions. To assess interpretability in practical applications, the KGQA benchmark is proposed to interpret the reasoning process in LLMs enhanced with KGs for open-ended, real-world question-answering scenarios [139]. LAMA [140] is a benchmark leveraged to test the ability of LMs to retrieve relations between entities and is used for LLM interpretability discussed in [141]. Another study used four benchmarks to systematically evaluate LLM interpretation with KGs in various domains[142]. The benchmark uses LAMA [140] and BioLAMA [143] to create factual questions, including two general-domain KGs: Google-RE [140], T-REx [144], and two domain-specific KGs: WikiBio [143] in the biology domain and ULMS [145] in the medical domain. Another approach using biomedical knowledge probing benchmark named MedLAMA for interpretability [146]. Furthermore, KagNet is another approach that uses a KG for commonsense reasoning, using CommonsenseQA[136]. [12] presents another study interpreting language models by integrating a KG using the benchmarks SQuAD [147] and Google-RE [140]. T-REx [144] is another benchmark used in language models to capture factual knowledge by integrating it with KGs [148]. Other popular benchmarks in approaches integrating KGs for interpretability in LLMs include WikiKG90M [149], Open Graph Benchmark (OGB) [150] and NELL-995. Question-answer-based benchmarks include WebQuestionsSP [151]  and FreebaseQA [152]. These benchmarks aim to evaluate various aspects of interpretability in KG-LLM integration, including reasoning paths, alignment between KG facts and LLM outputs, and the ability to handle complex, multi-hop queries across diverse domains.

\subsection{Benchmarks for Logic-integrated LLMs}
\label{sec6.2}
Datasets are essential tools for assessing the capabilities of LLMs, particularly in reasoning and explainability.  Researchers have developed various benchmark datasets to evaluate reasoning in symbolic integrated LLMs. This section explores multiple datasets used in symbolic integrated LLMs to highlight their role in addressing challenges in traditional LLMs, with a focus on reasoning. These datasets are categorised into complexity-level reasoning, domain-specific reasoning, and mode of reasoning. Table 8 presents benchmarks and evaluation metrics used in existing studies integrating logic with LLMs to perform reasoning.  

\subsubsection{Complexity-based reasoning benchmarks}
\label{sec6.2.1}

Reasoning datasets are categorised based on the complexity of steps designed to evaluate LLMs when integrated with symbolic AI. The benchmarks in this category focus on the level of complexity in steps while performing reasoning.  These datasets assess how well symbolic AI enhances LLMs' reasoning in existing studies. This includes single-hop reasoning and multi-hop reasoning. 

Single-Hop Reasoning: LLMs making direct inferences based on knowledge refer to single-hop reasoning. Various benchmarks include ProntoQA (FOL) [153], LogicNLI (FOL), and LogicBench (PL, FOL, NM) [154]. Multi-hop reasoning with Single Logic includes datasets for assessing LLMs’ reasoning by chaining through multiple steps to arrive at inference. This covers a single type of logic and only a few logical inference rules, including FOLIO (FL)[91], ProofWriter (rules-based logic—FOL) [155], SimpleLogic (PL), and PARARULE-Plus [156](Deductive). Existing reasoning datasets focus primarily on single-hop or multi-hop reasoning with limited inference rules. Multi-hop reasoning with non-monotonic and multiple logic covers datasets for drawing non-monotonic conclusions using multiple premises, various inference rules, and combinations. This includes multi-LogiEval [157], based on propositional, first-order, and non-monotonic reasoning. The design contains 30 inference rules and more than 60 combinations of basic inference rules. However, the complexity of reasoning depth needs further improvement. 

\subsubsection{Reasoning Modes}
\label{sec6.2.2}

This category explores existing benchmarks used in studies targeting symbolic integrated LLM. These benchmarks are essential for evaluating LLMs' performance across different reasoning models, such as inductive, deductive, and abductive reasoning. Deductive reasoning in symbolic AI-integrated LLMs refers to LLM’s ability to apply general rules to specific instances to deduce the conclusion.  This includes bAbi-deductive [158],  Winologic [159], WaNLI [160], Bigbech-deduction [161], Rulebert-Union [162], prOntoQA [153], ReClor[124],  LogiQA[163]. Inductive reasoning refers to deducing general conclusions from specific observations. Benchmarks in this category evaluate LLMs' ability to identify patterns to deduce generalised findings. This includes CLUTTR-Systematic[164], Bigbech-logical-args, bAbi-inductive. Abductive reasoning involves forming the most likely explanation for an observation. While less common in benchmarks, some datasets incorporate elements of abductive reasoning. These include LogiGLUE [165] and AbductionRule [166]. LOGIGLUE is a benchmark for testing LLM-integrated symbolic reasoning systems [165]. It contains 24 datasets for deductive, abductive, and inductive reasoning. LOGIGLUE instructs a language model named LogiT5 trained for single-task, multi-task, and chain-of-thought techniques to assess logical reasoning with limited external knowledge[171].

\subsubsection{Domain-Specific Reasoning}
\label{sec6.2.3}

Rapid improvements in approaches to integrating Symbolic AI with LLMs have enhanced domain-specific reasoning, resulting in new pathways for designing reasoning benchmark datasets specific to each domain. This section explores benchmark datasets that existing studies use to assess reasoning abilities in mathematical problems, coding, and algorithms.

Datasets for Mathematical and Physics Reasoning: Mathematical reasoning datasets assess LLMs’ capability to handle numerical and logical tasks. Benchmarks include Grade School Math (GSM) Symbolic[167], MAWPS [42]. These datasets assess LLMs’ problem-solving, arithmetic, and logical strengths, requiring sequential reasoning. ARB is another benchmark proposed to address advanced reasoning problems in multiple fields, including math and physics[168]. This benchmark provides quantitative analysis covering a high number of short answers and open responses compared to a low number of multiple-choice questions. Another linguistic physical reasoning benchmark in [45] introduces PiLoT (Physics in a Language of Thought), which enhances LLMs with symbolic reasoning by translating natural language into probabilistic programs that generate and simulate physical scenes. This framework enables structured logical inference about physical phenomena beyond simple pattern matching.

Datasets for Coding and Computational Reasoning: Programming and coding-related tasks use symbolic approaches to enhance LLM’s logical structure. Notable benchmarks used in existing studies include Abstraction and Reasoning Corpus (ARC) [169]and CodeXGLUE[170]. CodeXGLUE evaluates LLMs' ability to assess code intelligence across various programming-related tasks. ARC is a benchmark that can perform few-shot visual reasoning to determine an LLM agent’s ability to solve complex tasks.  
These benchmarks for symbolic-integrated LLMs across various domains highlight the potential to assess improved reasoning in mathematical, algorithmic, and coding tasks. 

\subsection{	Challenges and Considerations}
\label{sec6.3}

Data Contamination: Reasoning datasets used for training and instruction tuning may overlap with test examples used for evaluation, leading to inflated performance metrics that do not accurately reflect real-world reasoning capabilities.
Reliability of Metrics:  Standard metrics need to be improved to assess better the reasoning abilities of symbolic-integrated LLMs, particularly in logic-based tasks. More refined evaluation methods are required to process logical relationships and evaluate reasoning quality in LLMs that incorporate symbolic approaches.
The limited scope of Logic types: Real-world reasoning often involves a mix of deductive, inductive, and abductive reasoning, yet many benchmarks fail to cover these modes adequately. Additionally, these benchmarks tend to focus only on either PL or FOL, which limits the versatility of LLMs across diverse domains.
Scalability and complexity in benchmarks: The complexity of extensive and interconnected KGs challenges LLMs, making the tasks more computationally demanding.
Incompleteness and inconsistency: Benchmarks face issues of ambiguity and variability when translating natural language prompts into formal logic. Incomplete data in KGs can hinder reasoning, making it difficult for LLMs to fill knowledge gaps and retrieve facts reliably without relying on probabilistic assumptions.
Relevance and noise filtering: Another challenge involves filtering out irrelevant information while managing complex entity relations, which can introduce noise in multi-hop reasoning. 
Evaluation of interpretability: Development of diverse benchmarks is vital. Current benchmarks often focus on individual, specific reasoning types, limiting the scope of testing. They may not capture the full spectrum of symbolic reasoning across varying complexity levels and modes. Diverse datasets that encompass a broader range of reasoning modes and support complex, multi-step, multi-logic reasoning tasks should be developed. Benchmarks capable of measuring context maintenance across multi-hop KG relationships could improve integration, enhance KG alignment with LLMs, and optimise query efficiency during benchmark planning.
Domain-specific benchmarks are essential for designing and improving interpretability.
Hybrid reasoning and multimodality: For better real-world applicability, more comprehensive benchmarks that combine symbolic logic, KGS, and multimodal data are crucial.

\section{	Symbolic-LLM integration Role}
\label{sec7}

Integrating symbolic approaches with LLMs can address neural architectures' limitations while leveraging their strengths. This section discusses symbolic integration's role in enhancing LLM capabilities across various domains. Three critical areas explored in the section include knowledge representation with embeddings, reasoning, planning, and problem-solving.

\subsection{	Knowledge Representation and Embedding}
\label{sec7.1}

Symbolic integration with LLMs results in a well-structured representation of knowledge using embeddings. These embeddings translate logic and the entities and relationships of KGs into continuous vector spaces, capturing the semantic relationships between concepts. Improved semantic relationships can generate contextually relevant knowledge representations. KG-based embedding methods include translational distance and semantic matching models[172][173]. Advanced embedding techniques use a complex vector space to differentiate symmetric and asymmetric facts[174]. However, the limited structural connectivity in such approaches often results in incorrect handling of long-tailed relationships and does not enable the prediction of unseen entities[175]. Considering these challenges, current research is adapting LLMs for enhanced representations of entities and relationships[176]. A contrastive learning method named LMKE is proposed to address the problem of long-tail entities and improve the learning of embeddings generated by LLMs for KG[177]. LambdaKG is another LLM-based embedding approach proposed to enhance graph structure representation by concatenating neighbour entity tokens with the triple to feed the sentence into LLMs[178]. COMET is a framework proposed for dynamically expanding KGs that has demonstrated enhanced performance in reasoning tasks[179]. Another approach utilises GPT-3 to create high-quality knowledge triples to augment existing KGS [180], resulting in highly relevant knowledge retrieval.  Embedding symbolic knowledge into LLMs can result in high-quality domain-specific responses. KnowBert injects knowledge bases into language models to improve performance on entity-linked tasks[181]. Symbol-LLM injects symbolic knowledge into LLMs without compromising their general language understanding[67].

\subsection{Reasoning}
\label{sec7.2}

Symbolic reasoning refers to manipulating symbolic logic and KGs according to predefined rules and patterns to perform reasoning. LLM-guided symbolic reasoning bridges the gap between the statistical learning capability of LLM and the logic-driven process of the symbolic systems to gain state-of-the-art results on complex reasoning tasks. Symbolic systems excel in breaking down complex problems into step-by-step logical processes. Combining this strength with LLMs enables complex multi-hop reasoning over structured knowledge, such as knowledge graphs or rule-based systems[61]. Symbolic integration enhances logical consistency by integrating formal logic and symbolic solvers with LLMs.  A framework is proposed in [182] to generate intermediate reasoning steps using LLMs and step verification using symbolic solvers. ProofWriter uses LLM to create proofs verified by automated theorem provers and has achieved state-of-the-art results on several reasoning benchmarks [155]. Symbolic guided approaches are widely used for symbolic verification of LLM outputs to ensure reliability. LLM-generated outputs are verified using symbolic methods, and errors in task planning are reduced[183]. Such techniques are used to verify the factual accuracy of LLM-generated text using KGs[184]. KG-based RAG approaches can retrieve entities and relationships pertinent to queries, enabling the RAG system to perform deductive reasoning and coherent explanations [185]. Existing studies suggest that in LLM-integrated symbolic approaches, LLMs can act as symbolic reasoners capable of performing symbolic tasks in real-world applications [41]. While symbolic-integrated LLMs possess strong reasoning capabilities, the domain struggles to perform high-level multimodal reasoning. 

\subsection{Planning and Problem-solving}
\label{sec7.3}

LLM-integrated symbolic AI approaches can be utilised to solve complex planning systems. LLM-Planner is a system that employs LLMs to generate planners, refined by symbolic approaches to achieve significant improvement [104]. Plansformer uses LLMs to develop highly relevant symbolic plans [106]. Symbolic approaches have effectively solved complex problems by breaking them down into subproblems. Collaborative problem solving widely discussed in literature, emerges as a natural extension where multiple LLM-symbolic AI agents distributively tackle decomposed subproblems and coordinate their symbolic reasoning to achieve comprehensive solutions. Logic-LM proposed, and an LLM integrated, a symbolic approach using dedicated symbolic solvers to improve logical problem-solving on complex issues [59]. Nonetheless, intricate problem-solving strategies face difficulties arising from the token-level constraints inherent in LLM [75]. Incorporating symbolic reasoning techniques, such as chain-of-thought (CoT) and tree-of-thought (ToT), can enhance token-level problem-solving by encouraging a clear, step-by-step articulation of thought processes [75]. 

\subsection{Symbolic-integrated LLM to address explainability}
\label{sec7.4}

Explainability refers to the details and reasons a model provides to make its functioning and decisions clear. Different factors of responsible AI contribute to explainability, including reasoning, trustworthiness, reliability, robustness, safety, and security [186]. Various approaches are employed to address these challenges to ensure explainability. The explainability can exist at the data, model, and post-inference levels [18]. Data-level explainability involves using features, relevant examples, attention weights, or attributes from the dataset to generate an explanation for why an LLM made a particular prediction. Model-based explainability involves analysing the internal processes of the algorithm, focusing on LLMs' inner layers, embeddings, neurons, and activation functions to provide understanding. Post-inference explainability refers to interpreting the decision-making process after response generation, mainly for evaluating the LLM’s performance.
LLMs are the most robust transformer-based neural architecture capable of generating human-like content, but they present various challenges, with explainability being a notable concern. Table 9 offers a comparative analysis of different symbolic approaches integrated with LLMs, focusing on multiple aspects of explainability. The factors considered in this study include reasoning, robustness, and ethics. Table 9 provides a structured overview of this analysis. The comparison outlines how logic-based and graph-based integrations with LLMs like GPT and Llama function. It also highlights the limitations of each approach concerning specific explainability aspects and datasets, illuminating challenges researchers face. Common limitations include scalability issues and high dependence on the LLM used. The table also details the evaluation metrics employed to measure these aspects. By analysing the performance and limitations of each approach, this comparison demonstrates how different integration strategies—whether through KG, logic-based systems, or hybrid methods—impact the capability and scalability of LLMS. The focus lies on assessing the strengths and weaknesses of each process in terms of explainability. This comparison provides insights into the current state of symbolic integration and helps identify gaps in explainable, symbolic-integrated LLMs. These insights can serve as a foundation for future research development.  
\begin{table}[htbp]
\scriptsize
\caption{Benchmarks for KG-based Reasoning and Interpretability}\label{tab:domain-benchmark-metrics}
\renewcommand{\arraystretch}{1.2}
\begin{tabular*}{\textwidth}{@{\extracolsep\fill} p{1.5cm} p{7cm} p{3cm}}
\toprule
\textbf{Domain} & \textbf{Benchmarks} & \textbf{Evaluation Metrics} \\
\midrule
KGs for Reasoning &
GLUE [118], SuperGLUE [119], ATOMIC [122], XNLI [123], ReClor [124], HellaSwag [125], ComplexWebQuestions [126], MetaQA [127], GrailQA [128], CommonsenseQA [130], OpenbookQA [132], MedQA-USMLE [133], DocRED [138], SciTail [135] &
Accuracy, Reasoning, Recall, F1 score, Consistency, Precision, Reasoning steps, Reasoning depth \\
KGs for Interpretability &
LAMA [140], BioLAMA [143], Google-RE [140], T-REx [144], WikiBio [27], ULMS [145], CommonsenseQA [136], SQuAD [147], WikiKG90M [149], Open Graph Benchmark (OGB) [150], WebQuestionsSP [151], FreebaseQA [152] &
Completeness, Recall, Accuracy, Fidelity, Precision, Speed, Depth of Knowledge \\
\bottomrule
\end{tabular*}
\end{table}
\begin{table}[htbp]
\scriptsize
\caption{Logic-based Reasoning Benchmark and Evaluation Metrics}\label{tab:domain-sub-benchmark}
\renewcommand{\arraystretch}{1.2}
\begin{tabular*}{\textwidth}{@{\extracolsep\fill} p{1.5cm} p{2cm} p{5cm} p{3cm}}
\toprule
\textbf{Domain} & \textbf{Sub-domains} & \textbf{Benchmarks} & \textbf{Evaluation Metrics} \\
\midrule
\multirow{2}{2cm}{Reasoning based on Complexity} 
& Single-Hop Reasoning & ProntoQA (FOL) [153], LogicNLI (FOL), LogicBench (PL, FOL, NM) [154] & Accuracy, Precision, Recall, F1 score, Logical consistency, Reasoning depth \\
& Multi-Hop Reasoning & FOLIO (FL) [91], ProofWriter [155], PARARULE-Plus [156] (Deductive), Multi-LogiEval [157] & Accuracy, Consistency, Reasoning depth \\
\midrule
\multirow{3}{2cm}{Reasoning Modes} 
& Deductive & bAbi-deductive [158], Winologic [159], WaNLI [160], Bigbench-deduction [161], Rulebert-Union [162], PrOntoQA [153], ReClor [124], LogiQA [163] & Accuracy \\
& Inductive & CLUTTR-Systematic [164] & Accuracy \\
& Abductive & LogiGLUE [165], AbductionRule [166] & Accuracy, F1 score \\
\midrule
\multirow{3}{2cm}{Domain Specific} 
& Mathematics and Physics & GSM Symbolic [167], MAWPS [42], ARB [168], Linguistic Physical Reasoning Benchmark [45] & Accuracy \\
& Coding and Computation & Abstraction and Reasoning Corpus (ARC) [169], CodeXGLUE [170] & BLEU, CodeBLEU, Accuracy \\
\bottomrule
\end{tabular*}
\end{table}
\section{State-of-the-art Achievements and Challenges}
\label{sec8}

\subsection{State-of-the-art Achievements}
\label{sec8.1}

Integrating LLMs with symbolic AI has addressed various challenges, including enhanced reasoning, explainability, and benchmarks. Figure 09 demonstrates the state-of-the-art (SOTA) achievements and challenges. 

Methodological advancements considering architectural paradigms: In LLM to symbolic representation, various SOTA approaches discussed in the literature include symbolic-prompt engineering, symbolic guided instruction-tuned LLMs, schema and ontology-guided LLMs, Program of Thought, incorporating syntactic and semantic parsers and Tree-of-Thoughts. SOTA approaches for LLM-based action schema generation include few-shot prompting[104], corrective re-prompting[105], fine-tuning [106], agent planning [102], and hybrid LLM-symbolic methods [102] integrating predefined logic frameworks, schema libraries[58], and automated validation modules. In Symbolic to LLM architecture, innovative knowledge injection approaches include encoding symbolic rules into input prompts, embedding logic and KGs into LLMs, ontology-based injection and prompting, constraint-based fine-tuning, Logic-Informed Fine-Tuning with Symbolic Rules, Schema and Template-Based KG Injection, and hybrid NeSy approaches for validation or refinement. Predefined templates for symbolic structure conversion to natural language and domain-specific vocabulary-based context-aware translation are the latest breakthroughs for translation.

Innovative Approaches for Overcoming LLM Limitations: Integrating LLMs with symbolic AI has resulted in more interpretable and reasoning-oriented systems[187]. Symbolic AI influences the learning processes of LLMs, leading to enhanced decision-making. Large Language Model - Automated Reasoning Critic (LLM-ARC) is a NeSy framework that enhances the reasoning capabilities of LLMs by using Answer Set Programming (ASP) [188]. It has achieved state-of-the-art accuracy of 88.32\% on the FOLIO benchmark [187], [189]. Symbolic integration’s most significant achievement is improving the explainability of LLMs. Symbol-LLM [60] and SymbolicAI [33] are two approaches that use symbolic logic and rules to enhance explainability. Symbolic approaches are well known for learning from limited and unseen data, which can improve learning efficiency in LLMs[190][191]. This results in better knowledge transfer across domains without retraining the extensive models. Symbolic integration has resulted in enhanced problem-solving, as demonstrated by Logic-LM, which integrates LLMs with symbolic solvers to enhance logical problem-solving capabilities. This integration has improved LLMs' performance from various perspectives, but has increased parameter sizes and long processing times. 

Methodological advancements considering integration strategies: The integration of symbolic knowledge at various stages of LLMs has achieved state-of-the-art results. In the pretraining stage, integration is accomplished through instruction tuning, incorporating symbolic knowledge into model input, or embedding symbolic knowledge within training objectives, all considered breakthroughs. In the inference stage, cutting-edge approaches include symbolic prompting, probing, and retrieval-augmented knowledge fusion. During training and post-training integration, remarkable progress has been made using LLMs as symbolic encoders and decoders for constructing symbolic resources, including knowledge graphs and logic embeddings. In the post-training phase, leading innovations use symbolic structures to update the vast data already learned by LLMs without retraining the model. However, such approaches struggle with low-quality results, further compounded by noise from computational overhead. Strategies for effective knowledge injection remain an open research domain for discussion.

Knowledge extraction and validation from symbolic approaches overcome LLM’s challenges of incorporating external knowledge. The most state-of-the-art achievement is integration through RAG and parameter-efficient fine-tuning (PEFT) by encoding KG from node and relation perspectives using KG adapters[192]. Besides knowledge integration, knowledge validation using hybrid approaches and enhanced scalability, which improved accuracy and interpretability, are other significant progresses [193] [194]. The GSM8K benchmark is widely used to assess mathematical reasoning performance[195]. Based on GSM8K, GSM-Symbolic is designed to determine the variability in LLM performance across different formulations of the same problem, revealing limitations in logical reasoning capabilities [196]. SGP-Bench is another benchmark to evaluate LLMs' semantic understanding and consistency in symbolic graphics programs[197]. Further, Graph Neural Networks (GNNs) have continuously evolved and emerged as a robust framework for enhanced learning from graphs[198]. Cutting-edge approaches such as Graph Neural Networks (GNN)-enhanced LLMs and Graph Retrieval-Augmented Generation (GraphRAG) have emerged as pivotal innovations. GNN-enhanced LLMs leverage structured knowledge representations to refine contextual embeddings, allowing models to reason over relational data more effectively. 
Meanwhile, GraphRAG integrates graph-based retrieval mechanisms to ground LLM outputs in factual knowledge, reducing hallucinations and enhancing response reliability. Other notable breakthroughs include approaches like Joint Pre-training of Knowledge Graph and Language Understanding (JAKET), which simultaneously learns from textual and structured knowledge to improve adaptability across domains. These methodologies have demonstrated superior performance on benchmarks such as OpenBookQA and WebQuestionsSP, validating their effectiveness.


\subsection{Challenges and Future Directions}
\label{sec8.2}

Established Practices and Design Patterns: Current research lacks systematic examination of the design patterns that govern LLM-symbolic integration, which represents a substantial opportunity for theoretical advancement [92]. These practices require supporting evidence to validate the integration strategy. Similar proofs are necessary for selecting appropriate symbolic components, including first-order and higher-order logic, symbolic rules, and logical or probabilistic language. While existing approaches exhibit promising results[208], the absence of comprehensive pattern analysis limits the field's ability to develop coherent architectural principles. Future research should prioritize the identification and formalization of these patterns to enable more methodical and transferable approaches to NeSy system design.

Benchmarks: Another notable challenge is the need for standard benchmarks on symbolic knowledge and to ensure the quality and diversity of datasets through rigorous validation [206]. More diverse datasets suitable for different symbolic approaches are also regarded as necessary for evaluation [207].
Conflict resolution algorithms: A crucial challenge lies in designing conflict resolution algorithms that consistently resolve conflicting information between symbolic modules and LLM-generated responses to ensure consistency [208].

Hallucination detection: Symbolic-integrated LLM approaches have recently been employed to reduce AI hallucinations. Most literature on symbolic-integrated LLM approaches targets symbolic components as external knowledge bases to validate LLMs [209]. Some studies have proposed a generalised and seamless fact-checking model to detect hallucinations [210]. Research in symbolic AI for LLM hallucination detection requires advances in taxonomy development to classify hallucinations by their origin, such as those originating from symbolic AI knowledge bases versus LLM-generated errors, while analyzing contributing factors such as parametric knowledge conflicts and decoding strategies during integration [77]. In addition, uncertainty estimation techniques in sources and benchmarks require further research.

LLM knowledge alteration and fusion: Another challenge involves altering the knowledge within LLMs without retraining, utilising symbolic approaches. Some efforts are underway to modify LLM knowledge [211], but the literature also suggests this can cause cascading reactions affecting other relevant understanding [212]. Additionally, some LLMs only provide API access; hence, accessing their internal structure and parameters is further complicated. This complicates knowledge fusion during pre-training with symbolic data within the LLM pipeline [213]. Developing systematic frameworks for knowledge upgrading and addition in both open-source and closed-source LLMs is necessary.

Prompt-based reasoning: Current state-of-the-art prompting approaches are highly effective for extracting relevant LLM responses [214]. However, these methods may often require revision when assessing LLMs' relational reasoning capabilities using symbolic approaches [215].

RAG and evaluation metrics: The literature recognises RAG as a state-of-the-art achievement for refining LLM outputs without retraining [216]. Concerning symbolic AI, KG-based RAG is acknowledged as another significant success [217]; however, limited approaches target symbolic logic integration within the RAG pipeline. This gap highlights the need for further research to develop methods that specifically incorporate logic into RAG frameworks. Furthermore, evaluation metrics specific to graph-RAG and logic-RAG are considered an important area for future development, especially metrics emphasising reasoning and explainability.

Other challenges include developing mechanisms for incorporating real-time updates from dynamic KGs during training and fine-tuning [217]. Improving the integration of structured reasoning paths without compromising performance remains an area for advancement. LLMs face difficulties with graph linearisation and model optimisation, essential steps for input incorporation into LLMs [207]. Reducing model complexity and computational resource requirements, considering the efficiency of LLM and KG integration during training, is an ongoing research focus [218]. The literature contains limited investigations into specific adaptations of logic-based rules within the training phases of transformer architectures. Knowledge-infused attention and transformers require further active research. While symbolic integration enhances LLMs by producing more accurate and interpretable responses, addressing these challenges is vital to unlocking the full potential. 

\subsection{Conclusion}
\label{sec9}

This study examines the challenges associated with LLMs by addressing the requirements of high-risk sectors, where the explainability of generated responses is crucial to ensure transparency and interpretability. Various approaches are proposed to address these challenges, including NeSy approaches, which can enhance reasoning and logical thinking. NeSy AI combines neural networks with symbolic AI to improve prediction accuracy and interpretability. Based on the literature review and its robust capabilities, this survey explores NeSy AI as a suggested solution to address explainability. It discusses several studies showcasing distinct NeSy approaches applied to address various challenges of explainability in LLMs through symbolic techniques. The complementary nature of both fields has sharply increased the research trends. NeSy frameworks, specifically symbolic approaches, are widely discussed in the literature to ensure explainability in LLMs. However, current research overlooks applications for compiled integration of NeSy frameworks with LLMs, highlighting the need for further investigation. Furthermore, the practical applications of NeSyintegration with LLMs remain relatively rare, with most discussions still limited to theoretical frameworks.

Investigating architectural models that connect LLM-to-symbolic and symbolic-to-LLM pipelines reveals their potential in improving reasoning, explainability, and integration in AI systems. LLM-to-symbolic pipeline suffers from several challenges, including computational complexity, potential inconsistencies between LLM outputs and symbolic representations, dependency on large, annotated datasets, domain-specific adoption, sensitivity to prompt engineering, and limitations in scalability to complex symbolic structures. Researchers are exploring various solutions to address these issues, including parallel processing techniques, developing intermediate representation layers to bridge the gap between LLM outputs and symbolic inputs, and utilising adaptive weighting mechanisms in hybrid models. However, further research is needed to create more robust and efficient LLM-to-symbolic pipelines. A symbolic-to-natural language translation pipeline requires additional research in designing robust translation algorithms that handle various symbolic formats and provide context-aware translation. 
Furthermore, a significant gap that has yet to be explored comprehensively is the design of an efficient integration system to incorporate translated content into LLM prompts or processing pipelines, thereby enhancing LLM responses. In addition to the challenges mentioned earlier, future initiatives involving symbolic-to-LLM pipelines should concentrate on several key areas. These include enhancing integration at the symbolic contextual level and creating methodologies for verifying and validating symbolic knowledge. Focusing on these aspects can lead to significant improvements in symbolic-to-LLM pipelines. To summarise, while pipeline models provide some level of simplicity, they can struggle with complex, iterative reasoning, highlighting the need for hybrid models. 

Considering the reasoning capabilities of symbolic AI, a recommended direction explored in the literature is to combine the strengths of symbolic AI and LLMs for enhanced explainability. The paper discusses state-of-the-art achievements and challenges in integrating symbolic methods, such as KGs and logic-based reasoning, with LLMs at different lifecycle stages—pre-training, post-training, and inference. This article introduces stages for symbolic AI integration with LLMs, offering a structured approach to combining their strengths. The level of interaction between symbolic and LLM components is explored in terms of coupling. The algorithm-level and application-level perspective is analysed. A comparative analysis of existing approaches targeting explainability and symbolic-integrated LLMs is provided. This analysis can identify the insights of each approach to help understand which models offer better explainability, where improvements are needed, and how these systems perform specific tasks. Such an analysis provides valuable insights for future research and development in creating more transparent and reliable AI models. This review provides the research community with a clear roadmap for analysing challenges and achievements in symbolic-integrated LLM, fostering the development of new integration methods and promoting more systematic experimentation and innovation.

\begin{sidewaystable}[htbp]
\scriptsize
\caption{Comparative Analysis of Approaches Targeting Explainability}\label{tab:landscape_llm_symbolic}
\renewcommand{\arraystretch}{1.2} 
\begin{tabular*}{\textwidth}{@{\extracolsep\fill} p{0.8cm} p{2.5cm} p{3.2cm} p{1.3cm} p{2cm} p{2.5cm} p{2.5cm} p{2.5cm}}
\toprule
\textbf{Study} & \textbf{Explainability Factors} & \textbf{Approach} & \textbf{LLM} & \textbf{Symbolic} & \textbf{Dataset} & \textbf{Evaluation} & \textbf{Limitation} \\
\midrule
	2024 - [199] 	&	Ethics - Correctness and completeness ensured by ethical reasoning	&	Abductive-deductive framework combining LLMs with symbolic backward-chaining	&	GPT-3.5-turbo	&	Logic and rule based formal language through auto formalization (Prolog based)	&	ethical NLI benchmarks requiring commonsense reasoning [200]	&	macro-average f1 score at 3rd iteration for hard ethical statements is  58.6\% compared to 54. 1\% for CoT	&	Unable to provide satisfactory reasoning on complex ethical dilemmas, in different culture because of poor interpretation	\\
	2023 - [201]  	&	Robustness and reasoning	&	LLM integrated with logic programming named Answer set programming. Where LLM acts as semantic parser to give input for answer set programs. 	&	GPT-3,	&	Logic programming, Answer set Programming	&	Several NLP benchmarks, including bAbI, StepGame, CLUTRR, and gSCAN.	&	GPT-3(d3)+ASP on bAbI gives 99.99\% performance	&	Approach lacks handling errors in the dataset, difficult to write knowledge modules by hand.	\\

	2024 - [202]   	&	Faithful reasoning	&	Integrating LLMs with KGs to ensure interpretable reasoning by using structural information captured from KGs by considering relations paths	&	GPT, Alpaca, LLaMA2, and Flan-T5	&	KGs	&	WebQuestionSP (WebQSP) and Complex WebQuestions (CWQ)	&	For  WebQSP, Recall values after integrating with Rog planning for  GPT, Alpaca, LLaMA2, and Flan-T5 are  71.60,  74.20,  56.16 and  44.93	&	Highly dependent on planning, random plans gives even worst performance	\\
	2023 - [203] 	&	Robustness 	&	Tightly coupled LLM with cognitive architecture and symbolic rules to enhance the knowledge extraction and integration process 	&	GPT	&	Symbolic Rules	&	Prompts	&		&	Highly dependent on LLM layer	\\
	2023 - [204]  	&	Logical reasoning	&	LLM act as a semantic parser, to perform deductive inference integrated with FOL expressions	&	StarCoder+, GPT-3.5, GPT-4	&	First Order Logic to perform Logical reasoning, and deductive inferring	&	FOLIO and a balanced subset of ProofWriter 	&	26\% higher accuracy  with GPT-4 than CoT on ProofWriter (FOLIO)	&	Only one aspect of logical reasoning is focused, Scalability, computational costs, better reasoning and logic techniques needed	\\
\bottomrule
\end{tabular*}
\end{sidewaystable}

\section*{Declarations}

\begin{itemize}
    \item \textbf{Ethics approval and consent to participate:} This article is a review and does not involve human participants or animals; therefore, no ethical approval was required.
    
    \item \textbf{Consent for publication:} All authors have read and approved the final manuscript and consent to its publication.
    
    \item \textbf{Availability of data and material:} This article is a review and does not report new data. All sources used are cited in the References section.
    
    \item \textbf{Competing interests:} The authors declare that they have no competing interests.
    
    \item \textbf{Funding:} Not applicable.
    
    \item \textbf{Authors' contributions:} The authors' contributions are: \\ 
    Maneeha Rani: Conceptualization, Writing original draft, Editing. \\
    Bhupesh Kumar Mishra: Supervision, Proof-reading, Reviewing and Editing. \\
    Dhavalkumar Thakker: Supervision, Proof-reading, Reviewing and Editing.
    
    \item \textbf{Acknowledgements:} The first author acknowledges support from a PhD scholarship awarded by the University of Hull, UK. 
\end{itemize}

\begin{appendices}

\section*{Abbreviations} 

\scriptsize
\renewcommand{\arraystretch}{1.0} 

\begin{tabular}{p{2cm} p{10cm}} 
\toprule
\textbf{Abbreviation} & \textbf{Full Form} \\
\midrule
NeSy AI & Neurosymbolic Artificial Intelligence \\
ARC & Automated Reasoning Critic | Abstraction and Reasoning Corpus \\
ARB & Advanced Reasoning Benchmark \\
ASP & Answer Set Programming \\
ATOMIC & ATlas of MachIne Commonsense \\
ChatKBQA & Chat for Knowledge Base Question Answering \\
CodeXGLUE & Code-based General Language Understanding Evaluation Benchmark \\
CoK & Chain-of-Knowledge \\
CoLAKE & Contextualized Language and Knowledge Embedding \\
COMET & Commonsense Transformers for Automatic Knowledge Graph Construction \\
CoT & Chain of Thoughts \\
CWQ & Complex WebQuestions \\
DKPLM & Decomposable Knowledge-Enhanced Pre-trained Language Model \\
DocRED & Document-Level Relation Extraction Dataset \\
ERNIE & Enhanced Language Representation with Informative Entities \\
FOL & First Order Logic \\
FOLIO & First-Order Logic Inference \\
GAIN & Graph Aggregation-and-Inference Network \\
GLaM & Generalist Language Model \\
GLUE & General Language Understanding Evaluation Benchmark \\
GNN & Graph Neural Networks \\
Google-RE & Google-Relation Extraction \\
GSM8K & Grade School Math 8K \\
JointLK & Joint Reasoning with Language Models and Knowledge Graphs \\
K-BERT & Knowledge-enabled Bidirectional Encoder Representations from Transformers \\
Kagnet & Knowledge-Aware Graph Networks \\
KB-ANN & Knowledge-Based Artificial Neural Network \\
KELP & Knowledge Graph-Enhanced Large Language Models via Path Selection \\
KEPLER & Knowledge Embedding and Pre-trained LanguagE Representation \\
KGLM & Knowledge Graph in Language Models \\
KiL & Knowledge-infused Learning \\
Lama & LAnguage Model Analysis \\
LINC & Logical Inference via Neurosymbolic Computation \\
LLM & Large Language Model \\
LM & Language Model \\
LoT & Layer-of-Thoughts \\
MAWPS & Math Word Problems \\
MetaQA & Meta Question Answering \\
NM & Non-monotonic \\
PEFT & Parameter Efficient Fine-Tuning \\
PL & Propositional Logic \\
REALM & Retrieval-Augmented Language Model Pre-Training \\
ReClor & Reading Comprehension Dataset Requiring Logical Reasoning \\
RoG & Reasoning on Graphs \\
SGP-Bench & Symbolic Graphics Programs Bench \\
SQuAD & Stanford Question Answering Dataset \\
T-REx & Triples Based Relationship Extraction \\
TC–RAG & Turing-complete Retrieval-Augmented Generation \\
ToM-LM & Theory of Mind Reasoning in Language Models \\
WebQSP & WebQuestionSP \\
XNLI & Cross-lingual Natural Language Inference \\
LLM-ARC & Large Language Model - Automated Reasoning Critic \\
JAKET & Joint Pre-training of Knowledge Graph and Language Understanding \\
\bottomrule
\end{tabular}
\end{appendices}






\end{document}